\RequirePackage{snapshot}
\documentclass{article}

\usepackage{times}
\usepackage{graphicx}
\usepackage{sidecap}
\usepackage[small]{caption}
\usepackage{subcaption}
\usepackage{amsmath}
\allowdisplaybreaks
\usepackage{amsthm}

\usepackage{thmtools}
\usepackage{thm-restate}

\usepackage{inconsolata}

\usepackage{multicol}

\usepackage{setspace}

\usepackage{bm}

\renewcommand{\vec}[1]{{\boldsymbol{\mathbf{#1}}}}
\newcommand{\vw}{\vec{w}}
\newcommand{\vx}{\vec{x}}
\newcommand{\ve}{\vec{e}}
\newcommand{\vo}{\vec{o}}
\newcommand{\va}{\vec{a}}
\newcommand{\vq}{\vec{q}}
\newcommand{\vk}{\vec{k}}

\newcommand{\mQ}{\vec{Q}}
\newcommand{\mK}{\vec{K}}
\newcommand{\mV}{\vec{V}}
\newcommand{\calL}{\mathcal{L}}

\usepackage{pifont}
\usepackage{xspace}

\newcommand{\bos}{\texttt{\#}\xspace}
\newcommand{\eos}{\texttt{\$}\xspace}

\newcommand{\citeposs}[1]{\citeauthor{#1}'s \citep{#1}}

\usepackage{appendix}

\newcommand{\rightcomment}[1]{\(\triangleright\) {\small \it #1}}
\newcommand{\eqcomment}[1]{\addtocounter{equation}{1}\tag*{\rightcomment{#1}\quad(\theequation)}}
\usepackage{suffix}
\WithSuffix\newcommand\eqcomment*[1]{\tag*{\rightcomment{#1}}}

\usepackage{hyperref}

\usepackage[accepted]{icml2023}
\usepackage[textsize=tiny,disable]{todonotes}
\makeatletter
\newcommand*\iftodonotes{\if@todonotes@disabled\expandafter\@secondoftwo\else\expandafter\@firstoftwo\fi}
\makeatother
\newcommand{\noindentaftertodo}{\iftodonotes{\noindent}{}\ignorespaces}

\newcommand{\Fixme}[2][]{\noindentaftertodo}
\newcommand{\Notewho}[3][]{\noindentaftertodo}
\newcommand{\Jason}[2][]{\noindentaftertodo}
\newcommand{\Leo}[2][]{\noindentaftertodo}
\newcommand{\Hongyuan}[2][]{\noindentaftertodo}
\newlength{\extramargin}
\usepackage{calc}
\iftodonotes{\usepackage[paperwidth=\paperwidth+\extramargin*2+10mm,marginparwidth=\marginparwidth+\extramargin,width=\textwidth,height=\textheight]{geometry}}{}

\usepackage[utf8]{inputenc}
\usepackage[T1]{fontenc}
\usepackage{hyperref}
\usepackage{url}
\usepackage{booktabs}
\usepackage{amsfonts}
\usepackage{nicefrac}
\usepackage{microtype}
\usepackage{xcolor}

\usepackage{xpatch}

\usepackage{amsmath}
\usepackage{amssymb}
\usepackage{mathrsfs}
\usepackage{mathtools, cuted}
\usepackage{tabularx, booktabs}
\newcolumntype{C}{>{\centering\arraybackslash}X}
\newcolumntype{R}{>{\raggedleft\arraybackslash}X}
\newcolumntype{S}{>{\raggedleft\arraybackslash\hsize=.5\hsize}X}
\usepackage{latexsym}
\usepackage{url}
\usepackage{xspace}
\usepackage{bm,array}
\usepackage{amsfonts}
\usepackage{enumitem}
\usepackage{cases}
\usepackage{mathtools}
\usepackage{empheq}
\usepackage{bm}
\usepackage{esvect}
\usepackage[noabbrev,capitalize]{cleveref}
\crefname{equation}{Eq.}{Eqs.}
\crefname{footnote}{footnote}{footnotes}   
\crefname{line}{line}{lines}   
\crefname{assumption}{assumption}{assumptions}

\crefname{section}{\S}{\S\S}
\Crefname{section}{\S}{\S\S}
\crefformat{section}{#2\S#1#3}
\Crefformat{section}{#2\S#1#3}
\crefrangeformat{section}{\S\S#3#1#4--#5#2#6}
\Crefrangeformat{section}{\S\S#3#1#4--#5#2#6}
\crefmultiformat{section}{\S#2#1#3}{ and~\S#2#1#3}{, \S#2#1#3}{ and~\S#2#1#3}
\Crefmultiformat{section}{\S#2#1#3}{ and~\S#2#1#3}{, \S#2#1#3}{ and~\S#2#1#3}
\crefrangemultiformat{section}{\S\S#3#1#4--#5#2#6}{ and~\S\S#3#1#4--#5#2#6}{, \S\S#3#1#4--#5#2#6}{ and~\S\S#3#1#4--#5#2#6}
\Crefrangemultiformat{section}{\S\S#3#1#4--#5#2#6}{ and~\S\S#3#1#4--#5#2#6}{, \S\S#3#1#4--#5#2#6}{ and~\S\S#3#1#4--#5#2#6}
\usepackage{bbm}
\usepackage{float}

\let\frac=\tfrac

\usepackage[compact]{titlesec}

\newcommand{\defn}[1]{\textbf{#1}}

\renewcommand{\th}{\textsuperscript{th}\xspace}

\newcommand{\set}[1]{\mathcal{#1}}
\newcommand{\sat}{\textsc{Sat}\xspace}
\newcommand{\cnfsat}{\textsc{CNF-Sat}\xspace}
\newcommand{\sentence}{\vx\xspace}
\newcommand{\token}{x\xspace}
\newcommand{\setOfRollouts}{S}

\icmltitlerunning{Autoregressive Modeling with Lookahead Attention}

\begin{document}

\twocolumn[
\icmltitle{Autoregressive Modeling with Lookahead Attention}

\icmlsetsymbol{equal}{*}

\begin{icmlauthorlist}
\icmlauthor{Li Du}{jhu}
\icmlauthor{Hongyuan Mei}{ttic}
\icmlauthor{Jason Eisner}{jhu}

\end{icmlauthorlist}

\icmlaffiliation{jhu}{Johns Hopkins University}
\icmlaffiliation{ttic}{Toyota Technological Institute at Chicago}

\icmlcorrespondingauthor{Li Du}{leodu@cs.jhu.edu}
\icmlcorrespondingauthor{Hongyuan Mei}{hongyuan@ttic.edu}
\icmlcorrespondingauthor{Jason Eisner}{jason@cs.jhu.edu}

\icmlkeywords{Machine Learning, ICML}

\vskip 0.3in
]
\printAffiliationsAndNotice{}

\begin{abstract}
To predict the next token, autoregressive models ordinarily examine the past.  Could they also benefit from also examining hypothetical futures? 
We consider a novel Transformer-based autoregressive architecture that estimates the next-token distribution by extrapolating multiple continuations of the past, according to some proposal distribution, and attending to these extended strings.
This architecture draws insights from classical AI systems such as board game players: when making a local decision, a policy may benefit from exploring possible future trajectories and analyzing them. 
On multiple tasks including morphological inflection and Boolean satisfiability, our lookahead model is able to outperform the ordinary Transformer model of comparable size.
However, on some tasks, it appears to be benefiting from the extra computation without actually using the lookahead information.  
We discuss possible variant architectures as well as future speedups.

\end{abstract}

\section{Why Lookahead?}\label{sec:intro}

For some task-specific sequence distributions, the autoregressive modeling problem---guessing what word comes next---might intuitively benefit from considering words even farther in the future.  We propose an autoregressive architecture that looks ahead into the future.  While predicting the next token $\token_{t+1}$, our architecture model attends not only to the past tokens $\sentence_{\leq t}$ but also to a collection of ``lookahead strings'' $\sentence_{> t}$ sampled from some proposal sequence model.  This attention is used both at training time and at test time.

Lookahead is a staple of classical AI methods.  It is in principle necessary for fitting certain NP-hard sequence distributions in which the autoregressive conditional probabilities, being computationally intractable quantities, would otherwise require extremely large neural networks to model \citep{lin-etal-2021-limitations}.

However, those NP-hard distributions are artificial.  For naturally occurring sequences, why might one expect lookahead to help autoregressive modeling?  We argue that when the sequences represent an agent's behavior, an autoregressive parameterization is not always the \emph{simplest} description.  If the behavior is goal-directed---for example, an agent trying to achieve high reward in a Markov Decision Process---then the simplest description may include a characterization of the agent's environment and goals.  Even if the agent explicitly consults an autoregressive policy $p(\text{action} \mid \text{state})$ at each step, that policy is not arbitrary: while it may appear complex, it was shaped by reinforcement learning or by natural selection so as to achieve high-reward trajectories. That is, the simpler and deeper reason that the policy favors a particular action is that this raises the probability of obtaining a future reward.  Scientists are often able to produce simpler and more robust descriptions of human or animal behavior by couching them in terms of goals such as survival, reproductive advantage, information gain, social status enhancement, political strategy, and so forth.  In cognitive science, this idea is known as \emph{computational rationality} \citep{lewis-2014-computational}.

For example, autoregressive models of language are extremely popular \cite{radford2019language}.  Yet language is, of course, a goal-directed natural behavior that tends to successfully achieve communicative and other functions.  The high-probability word sequences may be (more or less) the ones that satisfy certain desiderata.  They are syntactically well-formed, achieve high harmony \cite{smolensky-1986}, satisfy poetic constraints such as rhyme \citep{ghazvininejad-etal-2017-hafez}, or achieve communicative goals \cite{Grice75,radford2019language}.  Thus, to choose the next word $\token_{t+1}$ given the left context $\sentence_{\leq t}$, a language model could benefit from also considering how the form and content of the sentence might evolve, for example by sampling plausible continuations $\sentence_{> t}$. It could then choose $\token_{t+1}$ that is likely given the desirable continuations $\sentence_{> t}$ but unlikely given the undesirable continuations. 

The NLP community has already developed models that look at the future, such as cloze language models $p(\token_t \mid \sentence_{<t}, \sentence_{>t})$ \citep{devlin-18-bert} as well as controllable generation models $p(\token_{t+1} \mid \sentence_{\leq t}, \text{desirable}(\sentence))$ \citep{yang-klein-2021-fudge}. The reinforcement learning community has also considered learning to condition actions on high future reward \citep{zhang-etal-2020-retrospective}.  However, none of these methods have examined samples of $\sentence_{> t}$ during training or inference.  In Monte Carlo Tree Search \cite{browne2012mcts-survey}, such samples are used during training, but not during inference, and there is no attention to the elements of $\sentence_{\leq t}$ but only an evaluation of $\text{desirable}(\sentence)$.

Of course, no matter how the true distribution $p(\sentence)$ over word sequences arose, it can always be factored autoregressively as $\prod_t p(\token_{t+1} \mid \sentence_{\leq t})$.  
Do autoregressive architectures lose anything by modeling it in this simpler way?  
The concern is that the local distributions $p(\token_{t+1} \mid \sentence_{\leq t})$ may be hard to learn from modest training data, or even to express as a tractable formula with a modest number of parameters. 
A speaker's next word (or more generally, an agent's next action) may simply be difficult to predict at a local level without a deeper understanding of what the entire utterance is designed to achieve.  When a reward-driven agent chooses $\token_{t+1}$ given $\sentence_{\leq t}$, it presumably does so either
\begin{itemize}
\item by explicitly planning ahead (``System 2'') to achieve high reward, or
\item by consulting a large precomputed lookup table or trained neural network (``System 1'') that essentially stores the results that would be obtained by planning ahead.
\end{itemize}
The terms ``System 1'' and ``System 2'' refer to \citeposs{kahneman2011thinking} notions of ``thinking fast and slow,'' respectively.  For example, one may play chess in either style: speed chess must use the intuitive System 1, whereas ordinary chess provides time for the deliberative System 2 to override the default behavior of System 1---in this case by evaluating the future consequences of System 1's proposals.

An autoregressive model of a rational agent will likewise need to adopt one of these strategies to correctly capture the agent's $p(\token_{t+1} \mid \sentence_{\leq t})$.  
Unfortunately, System 2's planning strategy is slow at test time, but System 1's precomputation strategy consumes space as well as training time.  Formalizing this predicament, \citet{lin-etal-2021-limitations} have pointed out that some weighted formal languages---even though they are defined by a $\mathrm{reward}(\sentence)$ function that is both fast to compute and concise to express---have autoregressive factors $p(\token_{t+1} \mid \sentence_{\leq t})$ that are NP-hard even to approximate, and thus computing or approximating them requires \emph{either} superpolynomial time \emph{or} superpolynomial space at test time, under commonly held complexity-theoretic assumptions.  

Perhaps the space requirements of the System 1 approach are tolerable in practice, and a sufficiently large neural network could do a reasonable job of approximating these autoregressive factors in the average case.  
Still, without a very large dataset, estimating the parameters of a very large network requires inductive bias (e.g., a prior) that encourages appropriate generalization.  
Specifically, the estimator should favor parameters such that the resulting autoregressive behavior $\prod_t p(\token_{t+1} \mid \sentence_{\leq t})$ admits a goal-directed explanation.  
Standard estimators such as $L^2$-regularized maximum likelihood do not have this property.
Instead, \citet{shi-etal-2018-diverse,mehta-etal-2020-irl} created inductive bias through inverse reinforcement learning, attempting to explain observed text by identifying a simple reward function along with a possibly complex policy that generates high-reward text.  
This policy then serves as the autoregressive language model.

A competing System 2 approach would drop autoregressive modeling altogether in favor of energy-based modeling \cite{Lecun06atutorial}.  
This is akin to learning the reward function without also learning a sequential generation policy.\footnote{Such a policy could be trained later by distilling the energy-based model into an autoregressive model---that is, compiling system 2 into system 1.  However, \citet{lin-etal-2021-limitations} caution that the autoregressive model may need to be much slower or much larger.}  
Instead, high-reward sequences are generated at runtime using an expensive planning-based process such as rejection sampling, MCMC, or stochastic beam search \citep{pmlr-v97-kool19a}.  
Linguistics famously made this move in the 1990's with the rise of Optimality Theory \citep{paradis-1988,prince-smolensky-2004}, which replaced complex stepwise generation procedures (akin to autoregressive models) with simpler direct descriptions of the rewards that those procedures were apparently constructed to obtain (akin to energy-based models).  Note that training energy-based models can be difficult, although noise-contrastive estimation is one approach.\footnote{One wrinkle is that when modeling strings of unbounded length, it is difficult to tell from the parameters of an energy-based model whether the distribution $p(\sentence)$ is well-defined, with a finite normalizing constant.}

In this paper, we attempt to find a practical hybrid approach in which System 2 consults System 1 \cite{kahneman2011thinking}.  We retain the autoregressive parameterization, but we allow the autoregressive factors to engage in a limited form of planning.  Specifically, our definition of $p(\token_{t+1} \mid \sentence_{\leq t})$ will use rollouts $p_0(\sentence_{> t} \mid \sentence_{\leq t})$ to consider the sentences that different choices of the next word $\token_t$ might lead to.  In this paper, we do not explicitly learn any reward function that evaluates the rollouts and chooses among them (although that is a reasonable direction for future work). Rather, we train a Transformer model $p$ (System 2) to predict the next word after freely examining rollouts from $p_0$ (System 1).

Our thinking is as follows.  Humans are able to speak in real time, so it is unlikely that they do exhaustive planning.  It is more plausible that they subconsciously engage in limited lookahead (bounded rationality).  If so, it is also plausible that their decisions can be influenced by all of the information in the lookahead scenarios that they consider, not only the reward, since this may compensate for the limited nature of the lookahead (i.e., there are only a few rollouts and they are off-policy).  If this is the case, then our architecture has a hope of being able to efficiently capture the behavior of human speakers.

We organize our paper as follows: in \cref{sec:model}, we introduce our lookahead architecture; in \cref{sec:experiment-overview}, we give an overview of the experimental tasks that we've chosen; finally, we discuss the details and results in \cref{sec:experiments} with additional ablation studies in \cref{sec:ablation}.
Our main contributions are as follows:
\begin{itemize}
    \item We present a novel model architecture that retains the autoregressive parameterization but enjoys a better estimate of the sequence distribution;
    \item We empirically demonstrate the effectiveness of lookahead models on several tasks where they outperform ordinary Transformer models with more parameters.  
    This suggests that shallow analysis of random futures predicted by the model can be comparable to deeper analysis of the observed past in some situations.
    \item We investigate the degree to which the lookahead mechanism actually exploits the lookahead strings.  Unfortunately, it is possible that for some tasks, the benefit comes from the extra computation and not the information revealed by lookahead. 
\end{itemize}
We close by outlining future directions for improving the model's predictive power, which we plan to investigate in a future version of this paper, as well as possible techniques for speeding it up. 

\begin{figure}[t]
    \centering
    \includegraphics[width=\linewidth]{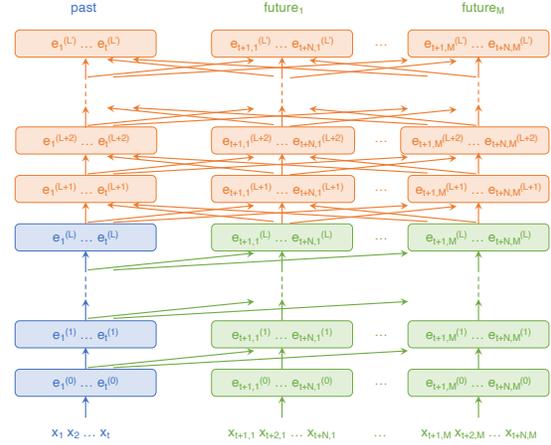}
    \caption{Our Transformer-based lookahead model.  The first column embeds the context string; each remaining column embeds a different lookahead string. 
    The blue blocks causally encode the prefix, and the parameters for this encoder are initialized to those of the base model $q$.  The green blocks causally encode the lookahead strings using the same parameters as the blue blocks.  Note that at these lower layers, each lookahead string is allowed to attend to the prefix but not vice versa. The orange blocks use bidirectional attention to further transform the token embeddings: at these higher layers, lookahead strings attend to one another and to the prefix in order to obtain embeddings that better predict $x_{t+1}$, as in the masked language model BERT \cite{devlin-18-bert}.}
    \label{fig:model}
\end{figure}

\section{Lookahead Transformer}\label{sec:model}
An autoregressive sequence model defines the probability of any sequence of symbols \mbox{$\sentence_{1:T} =\token_1 \ldots \token_{T}$} by a product of conditional probabilities of the symbols, as shown in \cref{eq:autoregressive}. Here $\sentence_{s:t}$ denotes the substring $x_{s} \ldots \token_{t}$.
In this paper, we propose a technique for enhancing autoregressive sequence modeling with \emph{lookahead}. 
Technically, we model the probability of a sequence as%
\begin{subequations}
\begin{align}
    p(\sentence_{1:T})
    &=\prod_{t=0}^{T-1} p( \token_{t+1}\mid \sentence_{1:t} ) \label{eq:autoregressive} \\
    &= \prod_{t=0}^{T-1} \sum_{\setOfRollouts_t} q(\setOfRollouts_t \mid \sentence_{1:t}) p( \token_{t+1} \mid \sentence_{1:t}, \setOfRollouts_t ) \label{eq:expectation}
\end{align}\label{eq:ar}
\end{subequations}
where $\setOfRollouts_t$
is a collection of $M$ \defn{lookahead strings} of length $N$ that are rolled out from a base model $q$, which we take to be a pretrained autoregressive model:
\begin{subequations}
\begin{align}
    \setOfRollouts_t &= \{ \sentence_{t+1:t+N,1}, \ldots, \sentence_{t+1:t+N,M} \} \\
    \sentence_{t+1:t+N,m} &\sim q( \cdot \mid \sentence_{1:t} ) \text{ IID for $m=1,\ldots,M$}
\end{align}\label{eq:samples}
\end{subequations}
In practice, we stochastically approximate the expectation \labelcref{eq:ar} by using \labelcref{eq:samples} to sample a single set $\setOfRollouts_t$ of size $M$ at each $t$.\footnote{Models that give stochastic probability estimates are not common, but may be familiar from dropout \citep{srivastava14a}.  Unlike the dropout literature, we use these stochastic estimates even in evaluation, since the expectation is intractable.}
Given these samples, \cref{eq:ar} is approximated by
\begin{align}
    p(\sentence_{1:T})
        &= \prod_{t=0}^{T-1} p( \token_{t+1} \mid \sentence_{1:t}, \setOfRollouts_t ) 
\end{align}

In practice, we take $q$ to be a Transformer sequence model---one that has been pre-trained to predict the next word without any lookhead.  As $q$ is presumably imperfect, we hope to improve on it with our lookahead model $p$.  We propose the following Transformer-based architecture for $p$, which is illustrated in \cref{fig:model}.
At each $t$:%
\begin{enumerate}
    \item For each $m=1,\ldots,M$, define $\sentence_{1:t+N,m}$ to be $\sentence_{1:t}\, \sentence_{(t+1):(t+N),\,m}$, which concatenates the observed past $\sentence_{1:t}$ with the $m$\textsuperscript{th} lookahead string.
    \item For each $m$, embed the tokens of $\sentence_{1:(t+N),m}$ using $L$ Transformer layers, giving a sequence of $t+N$ vectors, $\ve^{(L)}_{1:(t+N),m}$. 
    The attention used in these layers is \emph{causal} (see \cref{sec:causal}). In other words, the observations $\sentence_{1:t}$ are not allowed to attend to the lookahead strings (so their embeddings do not vary with $m$), and the lookahead strings are not allowed to attend to each other.\looseness=-1 
    \item Further transform the collection of $t+MN$ embedded tokens $\ve^{(L)}_{1:(t+N),m}$ using additional Transformer layers $\ell \in \{L+1, \ldots, L'\}$. 
    The attention in these upper layers is \emph{bidirectional} (see \cref{sec:bidirec}): as a result, the observations $\sentence_{1:t}$ are now allowed to look at the lookahead strings, and the lookahead strings are now allowed to look at each other. 
    \item 
    Define
    \begin{align}\label{eq:pdef}
        p( \token_{t+1} &= v \mid \sentence_{1:t}, \setOfRollouts_t ) \propto \exp(\vo_v^\top \ve^{(L')}_{t})
    \end{align}
    where  $\set{V}$ is the finite vocabulary and $\vec{o}_{v}$ is a learned embedding for each word $v \in \set{V}$.
\end{enumerate}

\subsection{Token Embeddings (layer $0$)}

Where $s$ is a token position in the $m$\textsuperscript{th} concatenated string, the token's layer-$0$ embedding $\ve^{(0)}_{s,m}$ is determined by its type $v = x_{s,m} \in \set{V}$, together with its position $s$: $\ve^{(0)}_{s,m} = \vec{o}_v + \vec{p}(s)$ where 
$\vec{p}(s)$ is the sinusoidal positional encoding used by \citet{vaswani-2017-transformer}.
Note that the positional encoding does not depend on which lookahead string we are embedding (i.e., $\vec{p}(s)$ does not depend on $m$). 

\subsection{Causal Lookahead Attention (layers $1$ to $L$)}\label{sec:causal}

The goal of our lower $L$ layers is to embed each of the concatenated strings separately, just as a standard Transformer decoder would.  There is no communication among the strings.  In other words, each concatenated string $\vx_{1:t+N,m}$ is transformed to a sequence of multidimensional vectors $\ve^{(L)}_{1:t+N,m}$.   Intuitively, in the framing of \cref{sec:intro}, these lower layers extract the properties of $\vx_{1:t+N,m}$ that allow the model to evaluate whether this rollout yielded a high-reward sequence.%

Each layer $\ell\in\{1,2, \ldots,L\}$ resembles a generative Transformer layer~\citep{vaswani-2017-transformer,radford2019language}. 
The embedding $\ve^{(\ell)}_{s}$ of the $s$\th token in $\vx_{1:t+N,m}$ is given by 
\begin{align}
    \vec{e}^{(\ell)}_{s,m} 
    &= f^{(\ell)}\left( \vw^{(\ell)}\, \left[\vec{e}^{(\ell)}_{s,m,1}; \dotsm; \vec{e}^{(\ell)}_{s,m,H} \right] + \vec{e}^{(\ell-1)}_{s,m} \right)
\end{align}
where $f^{(\ell)}(\ve) = \text{FFN}^{(\ell)}(\ve) + \ve$ and $\text{FFN}^{(\ell)}$ is a feed-forward network with $\ell$-specific parameters. 

Each $\ve^{(\ell)}_{s,m,h}$ is given by a causal attention head that looks at the layer $\ell-1$ embeddings of its left context along with itself, i.e., $\vec{e}^{(\ell-1)}_{1:s,m}$. 
Concretely, it is
\begin{subequations}\label{eq:causal-attn}
\begin{align}
    \ve^{(\ell)}_{s,m,h}
    = 
    \sum_{r=1}^{s}  \alpha^{(\ell)}_{s,r,m,h} \vec{v}^{(\ell)}_{r,m,h}, \\
    \alpha^{(\ell)}_{s,r,m,h} 
    \propto \exp
    \left(
    \frac{1}{\sqrt{D}} \vec{k}^{(\ell)}_{s,m,h}\phantom{}^{\top} \vec{q}^{(\ell)}_{r,m,h} 
    \right)
\end{align}
\end{subequations}
where $D$ is the dimension of the embedding. The $\vec{q}^{(\ell)}_{r,m,h}$ $\vec{k}^{(\ell)}_{r,m,h}$ $\vec{v}^{(\ell)}_{r,m,h}$ are the \defn{query}, \defn{key}, and \defn{value} vectors. They are defined by:
\begin{subequations}
\begin{align}
    \vq_{r,m,h}^{(\ell)} = \mQ^{(\ell)} \ve_{r,m,h}^{(\ell-1)}, \\ 
    \vk_{r,m,h}^{(\ell)} = \mK^{(\ell)} \ve_{r,m,h}^{(\ell-1)}, \\
    \vec{v}_{r,m,h}^{(\ell)} = \mV^{(\ell)} \ve_{r,m,h}^{(\ell-1)}.
\end{align}
\end{subequations}

Since all $M$ of the concatenated strings share a prefix $\vx_{1:t,m}$ that is embedded using causal attention that does not depend on the suffix, the resulting embedding $\ve^{(\ell)}_{1:t,m}$ (for any $\ell \in [0,L]$) is independent of $m$.  Thus, only one copy of it is shown in \cref{fig:model}.  

\subsection{Bidirectional Lookahead Attention (layers $L+1$ to $L'$)} \label{sec:bidirec} \label{sec:lookahead}

When constructing higher layers, our attention becomes unrestricted.  We continue to use only a single shared embedding of the prefix, but we now use bidirectional (non-causal) attention, so it can attend over all of the suffixes. The suffixes can attend to the shared prefix and to one another.  
Intuitively, these higher layers examine the ensemble of rollouts in order to predict the next symbol.  This is beneficial if, as discussed in \cref{sec:intro}, the \emph{true} generative mechanism also generates the next symbol by using some kind of planning ahead, e.g., trying to achieve a high-reward sequence.

The head dimension $h$ is omitted in the equation below for simplicity.
\begin{align}
    \ve_{s,m}^{(\ell)} = 
        \sum_{r=1}^t \alpha_{s,r,m, m}^{(\ell)}\vec{v}_{r,m}^{(\ell)} + 
        \sum_{m'=1}^M \sum_{r=t+1}^{t+N} \alpha_{s,r,m,m'}^{(\ell)} \vec{v}^{(\ell)}_{r,m'}, \label{eq:bidirect-attn} 
\end{align}
where
\begin{align}
    \alpha_{s,r,m,m'}^{(\ell)} \propto 
    \begin{cases} 
        0 & \hspace{-1.8cm}\text{$m'\not=m$ \& $1\leq r \leq t$,} \\
        \exp\left(\frac{1}{\sqrt{D}}\vk_{s,m}^{(\ell)}\phantom{}^\top \vq_{r,m'}^{(\ell)}\right)  & \text{otherwise}.
    \end{cases}
\end{align}
In contrast to \cref{eq:causal-attn}, \cref{eq:bidirect-attn} allows every token to attend to every other token---the $t$ tokens in the prefix $\sentence_{1:t}$ and also the $N\times M$ tokens in all of the hypothetical future roll-outs $\sentence_{t+1:t+N,m}$.

Finally, we use \cref{eq:pdef} to model the autoregressive distribution of the next symbol $\token_{t+1}$ given $\sentence_{1:t}$.

\begin{table}[t]
    \centering
    \begin{tabular}{lcc}
    \toprule
         & Avg. Loss ($\downarrow$) & Avg. Acc ($\uparrow$) \\
    \midrule
       3-Layer Baseline  & 0.489 & 87.5 \\
       4-Layer Baseline  & 0.486 & 87.8 \\
       5-Layer Baseline  & {\bf 0.476} & {\bf 88.6} \\
    \midrule
       (3+1)-Layer Lookahead & {\bf 0.476} & {\bf 88.9} \\
       (3+2)-Layer Lookahead & {\bf 0.476} & {\bf 88.9} \\
    \bottomrule
    \end{tabular}
    \label{tab:sat-results}
    \caption{Test loss (log-loss per token) and argmax-prediction accuracy averaged over 50 randomly sampled 3-\sat formulas for Lookahead Transformer and baseline Transformers.  We boldface the best (lowest) result and all others that are not significantly worse (paired permutation test by formula, $p<0.05$).  
    }
\end{table}

\begin{figure}[t]
    \centering
    \includegraphics[width=0.48\textwidth]{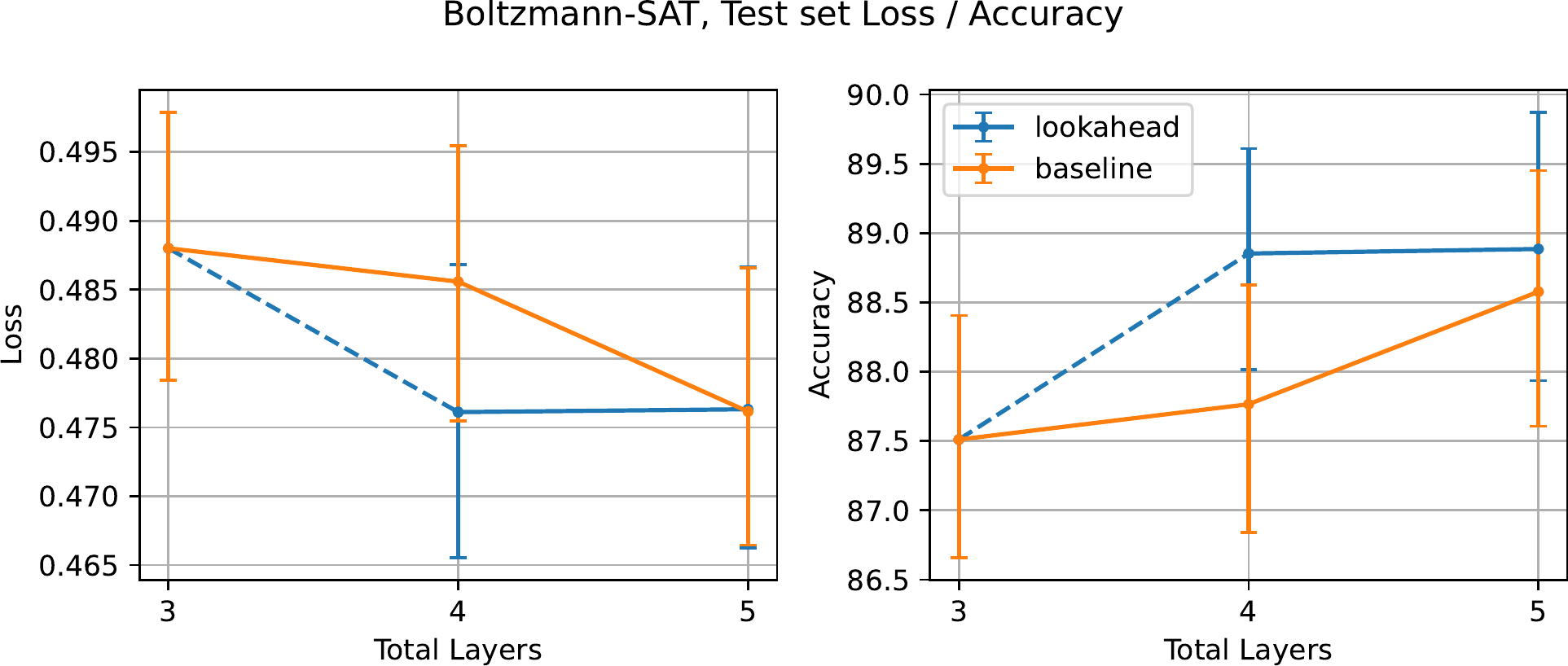}
    \caption{Mean loss over 50 formulas with error bars showing a 95\% bootstrap confidence interval. 
    The error bars are wide because the formulas vary in difficulty, but for any given formula, the lookahead method tends to do better, which is why it achieves statistical significance in a paired test (\cref{tab:sat-results}).}
    \label{fig:sat-test-results}
\end{figure}

\section{Experimental Tasks} \label{sec:experiment-overview}

We first describe and motivate our experimental tasks at a high level, reserving full details for the experimental section (\cref{sec:experiments}).

\subsection{Learning Boltzmann Distributions} \label{sec:boltzmann-sat-intro}

As pointed out by \citet{lin-etal-2021-limitations}, when a sequence distribution is defined with reference to an NP-complete problem such as Boolean Satisfiability (\sat), it can be easy to determine the relative probabilities of complete sequences, but intractable to compute the autoregressive probabilities $p(x_t \mid \vx_{<t})$. This is because for NP-complete problems, verifying a witness is fast but finding or completing a witness is slow.  Classical algorithms for finding witnesses in \sat (satisfying assignments) rely on backtracking search, which is an exhaustive form of lookahead \cite{dpll-1962}.  We therefore hope that stochastic lookahead may provide at least some benefit to autoregressive models.

We experiment with the natural Boltzmann distributions induced from \cnfsat instances.  Let $\phi$ be a Boolean formula with $n$ variables given in CNF, e.g., $\phi=(x_0\lor x_2 \lor \lnot x_3) \land (x_2 \lor x_3 \lor \lnot x_4)$, so that a bit string $\vx = x_0 x_1 \ldots x_{n-1} \in \{\texttt{0},\texttt{1}\}^n$ represents a possible assignment to the variables of $\phi$.  We define the Boltzmann distribution induced by $\phi$ at temperature $T > 0$ as the following:
\begin{subequations}
\begin{align}
E(\vx)&=\#\text{\{clauses in } \phi \text{ violated by } \vx  \}, \\
Z_T&=\sum_\vx \exp(-E(\vx)/T), \\
p(\vx)&=\frac{\exp(-E(\vx)/T)}{Z_T}.
\end{align}
\end{subequations}

Such a distribution admits a very short description, namely the formula $\phi$, which could be used to optimally predict the next symbol using \emph{exhaustive} lookahead.  While our neural lookahead architecture is parameterized by real numbers rather than by a formula, and does not do exhaustive lookahead, our hope is that it can still produce a comparatively compact approximation to the distribution. 
In other words, we hope that it allows us to fit the distribution using fewer parameters than a baseline architecture, and thus generalize from less training data.

To ensure that we get relatively hard instances of \sat, we take advantage of the well-studied family of random $k$-\sat problems \citep{annals.2022.196.1.1}.
A random $k$-\sat instance consists of $n$ variables and $m$ random clauses.
Each of the $m$ clauses is independently sampled from a uniform distribution over all $\binom{n}{k}2^k$ possible clauses. $\alpha=\frac{m}{n}$ is defined as the \emph{clause density}.
For each $k \geq 3$, there is a \emph{phase transition threshold} $\hat\alpha_k$ such that as $\alpha$ increases past a threshold number $\hat\alpha_k$ (keeping $n$ fixed) the probability of a randomly sampled instance being satisfiable quickly drops from 1 to 0. It has been widely observed that \sat solvers are slow to determine the satisfiability of a formula when $\alpha$ is near $\hat\alpha_k$ \citep{cheeseman-1991,mezard-etal-2002}.
Therefore, as a first test-bed to investigate whether lookahead can improve autoregressive models where the true autoregressive probabilities are expensive to determine, we draw our training data from Boltzmann distributions induced from random 3-\sat instances where the clause density is close to the threshold, i.e. $\frac{m}{n}\approx \hat\alpha_3\approx 4.259$.  We give more details in \cref{sec:sat-results}.

\subsection{Letter Infilling} \label{sec:letter-infill-intro}

Lookahead and planning are integral to many generative language tasks when modeled autoregressively. For example, generating poetry autoregressively involves carefully selecting words based not only on whether their own syntax, meaning, rhyme, and meter fit the constraints of poetic language, but also on whether they provide opportunities for \emph{future} words to fit those constraints.
Hence, planning for future events is necessary, as both the semantic content and phonological structure of the words are important to the success of poem writing \citep{ghazvininejad-etal-2017-hafez}.  For example, if lookahead finds that ``tower'' would be a good rhyming word to end the current line, we might want to bias generation of the earlier words in the line so that they are semantically and syntactically related to ``tower.''

As an initial exploration of this space, we designed a simpler NLP task where lookahead could be beneficial, namely infilling missing letters in a word. Specifically, given a word that was never seen during training, such as \texttt{d i s c r e p a n c y}, we randomly mask out some characters to get a form like \texttt{d - s - - e - a - c -}, and task the model with generating the original form. Intuitively, rollouts will help the model make a more informed choice of each letter by considering multiple ways in which different choices might play out to produce a well-formed novel word that is consistent with the constraints.

Our task is subtly different from \emph{controlled generation} \citep{zhang-etal-2022-survey}, which must also anticipate future events.  Controlled generation attempts to generate text that has specified attributes or a high reward .  For example, FUDGE \citep{yang-klein-2021-fudge} and particle smoothing \citep{lin-eisner-2018-naacl} are both left-to-right string decoders which, at each step, consult an internal model to assess whether the prefix generated so far is likely to extend into a high-scoring result.  It would be interesting to enhance those internal models to inspect lookahead strings.  However, in those methods, the internal models are used only to guide search with respect to a fixed and given $p(\vec{x})$.  In our task, by contrast, we are still training $p(\vec{x})$ (to maximize likelihood) and the lookahead is incorporated into its actual definition.

\subsection{Morphological Inflection} \label{sec:morph-intro}

The previous tasks were artificial, so we now turn to a more realistic task, where it is less clear \emph{a priori} whether lookahead will help in practice.

Morphological inflection is a common NLP task \citep{cotterell-etal-2017-conll} where the model is given a lemma (such as ``\texttt{run}'') and a set of morphological tags (such as \verb|{verb,participle,past}|), and is
required to generate the inflected word form (in this case, ``\texttt{running}'').  

Inflection often involves concatenating prefixes and/or suffixes to the lemma, but this concatenation may trigger other changes to pronunciation or spelling (such as the doubled ``\texttt{n}'' in ``\texttt{running}''). 

Linguists have found that the simplest descriptions of these changes involve multiple left-to-right and right-to-left editing passes \cite{kenstowicz-1993}, or better yet, global discrete optimization \cite{prince-smolensky-2004}.  A strictly left-to-right autoregressive model would presumably have to be more complex to explain the same patterns.  That is, it would require more parameters and more training data to capture the $x_t$ that would be predicted by the linguists' mechanisms.  Since the linguists' mechanisms involve interactions between $x_t$ and $\vx_{>t}$, it is possible that looking ahead to possible $\vx_{>t}$ values will make it easier for an autoregressive model to predict $x_t$.

\begin{figure*}[t]
    \centering
    \includegraphics[width=0.49\textwidth]{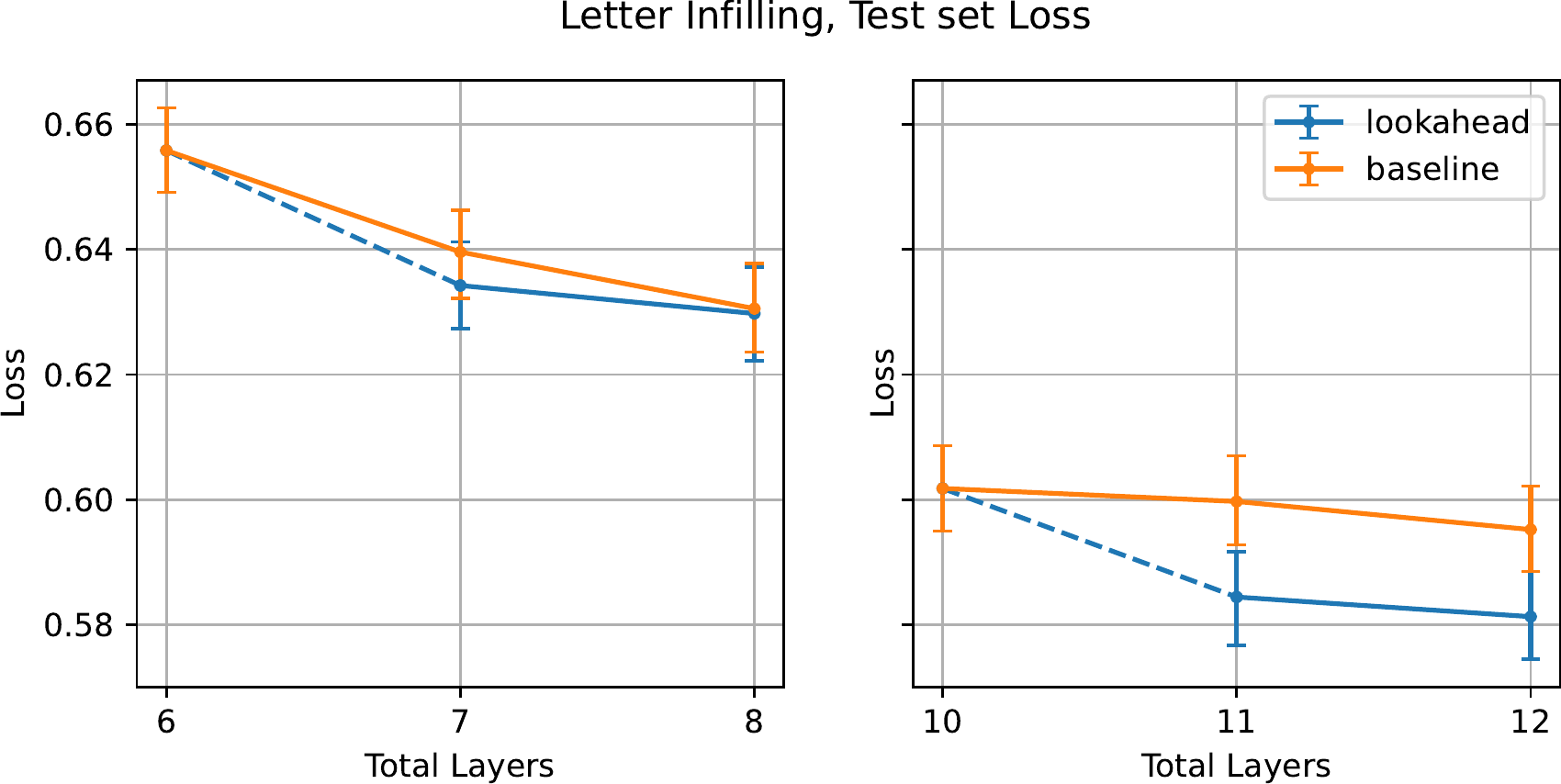}
    \includegraphics[width=0.49\textwidth]{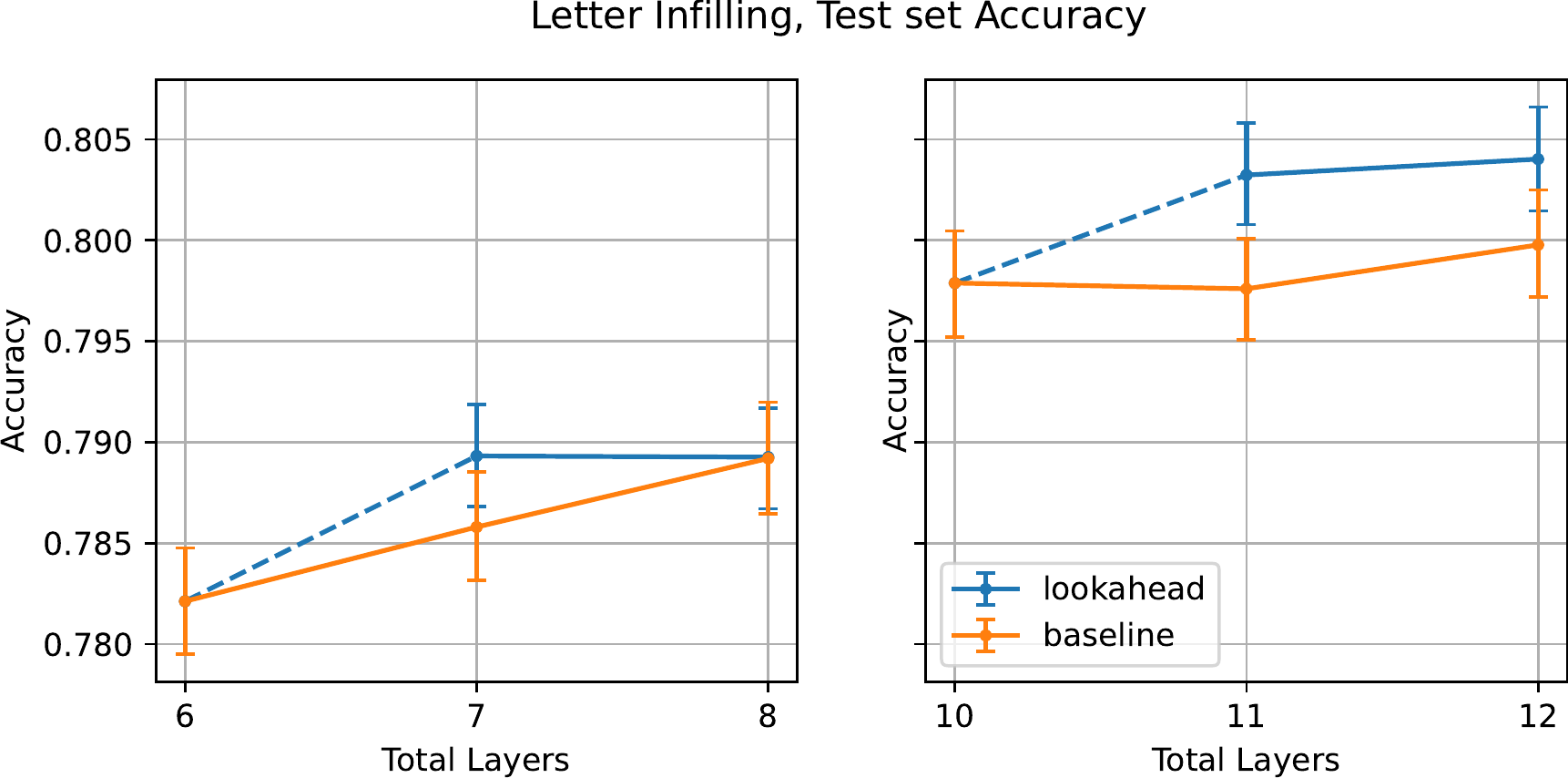}
    \caption{Test set statistics of Lookahead Transformers and baseline Transformers with different number of layers for the letter infilling task (with $95\%$ error bars computed using bootstrap resampling). In each plot, the blue line shows the result of adding 0, 1, or 2 bidirectional lookahead attention layers to the 6 or 10 base causal layers, whereas the orange baseline shows the result of adding 0, 1, or 2 ordinary causal layers.  All points are trained for the same total number of epochs.}

    \label{fig:synth-test-results}
\end{figure*}

\begin{figure*}[t]
    \centering
    \includegraphics[width=0.49\textwidth]{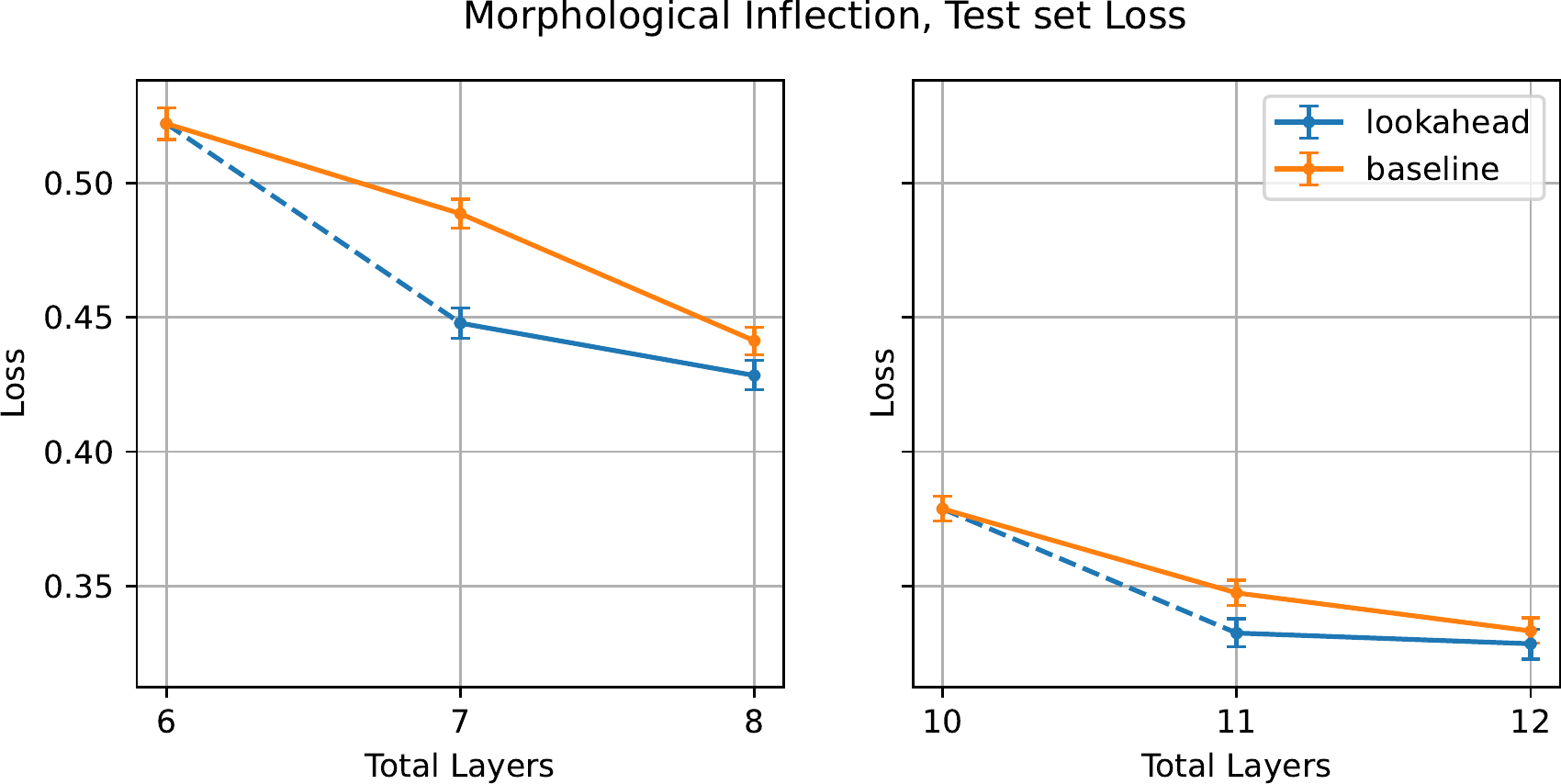}
    \includegraphics[width=0.49\textwidth]{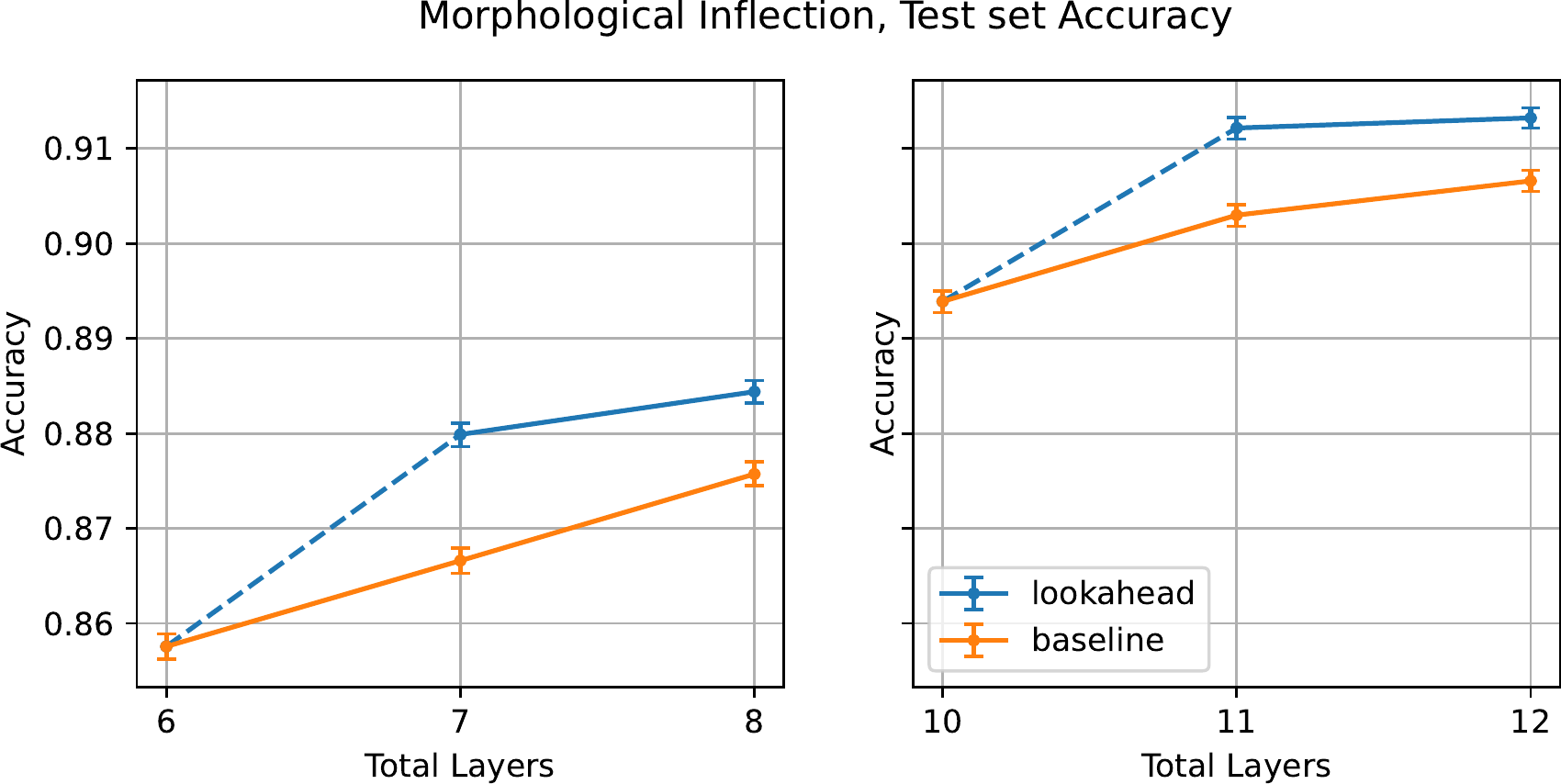}
    \caption{Test set statistics of Lookahead Transformers and baseline Transformers with different number of layers for the morphological inflection task (with $95\%$ error bars computed using bootstrap resampling). 
    In each plot, the blue line shows the result of adding 0, 1, or 2 bidirectional lookahead attention layers to the 6 or 10 base causal layers, whereas the orange baseline shows the result of adding 0, 1, or 2 ordinary causal layers.  All points are trained for the same total number of epochs.}
    \label{fig:morph-test-results}
\end{figure*}

\section{Experimental Setup and Results} \label{sec:experiments}

We evaluate our lookahead framework in several simulated and real-world datasets with different magnitudes of vocabulary size and training dataset size, and compare our models against Transformer baselines.

\paragraph{Sequence Model} Each training and test example has the form $(\vec{x},\vec{y})$.  Our model is a Transformer decoder, as commonly used in prompted language modeling \cite{brown-2020-gpt}.  When prompted with the context $\vec{x}$ followed by the special symbol \bos, the model is trained to maximize the log-probability of continuing it with the desired output sequence $\vec{y}$ followed by the special symbol \eos.  
Thus the training loss on the example $(\vec{x},\vec{y})$ is given by
\begin{align*}
    \mathcal{L}_\text{seq2seq} = \sum_{t=S}^{T}
    -\log p\left(z_{t+1} \mid \vec{z}_{1:t}\right).
\end{align*}
where $\vec{z} = \vec{x} \bos \vec{y} \eos$, $S = |\vec{x} \bos|$, and $T=| \vec{x} \bos \vec{y}|$.

As usual in this architecture, each token is embedded (transformed) using the same parameters regardless of whether it falls in $\vec{x}$ or $\vec{y}$, and is embedded using
attention only to itself and preceding tokens.  In particular, a token of $\vec{x}$ cannot attend to later tokens in $\vec{x}$, as it could if a separate encoder were used, nor can it attend to tokens in $\vec{y}$.

\paragraph{Experimental Setup} Here, we describe the common training and evaluation setup in all our experiments. For each dataset, we first train a baseline $L$-layer Transformer where $L$ is set to 6 or 10 in two of our three experiments. To train a lookahead model with $L'$ layers (of which $L$ layers are causal attention layers and $L'-L$ layers are bidirectional lookahead attention layers), we use the trained $L$-layer Transformer both as the proposal distribution $q(\cdot\mid\vx_{1:t})$ and to initialize the $L$ causal layers of the lookahead model.
The remaining $L'-L$ lookahead layers are randomly initialized.  We only consider $L' = L+1$ and $L' = L+2$.  (We did not try $L+3$ since $L+2$ turned out not to improve much over $L+1$.)\looseness=-1

Due to the Lookahead Transformer being well-initialized at its lower layers, all our experiments train the Lookahead Transformers with 20\% of the number of steps used to train baseline models. For example, if we have trained our baseline Transformer for 100 epochs, we will train the corresponding Lookahead Transformer for only 20 epochs.

As a baseline, we separately train an $L'$-layer Transformer, so that we can compare our lookahead model to a simpler and faster model with the same number of parameters.

For all our experiments, we train with Adam \citep{kingma-15} and apply dropout with probability 0.1. We also fix the number of lookahead strings to be $M=5$ and the length of lookahead strings to be $N=5$.

\subsection{Learning Boltzmann Distributions}\label{sec:sat-results}

In \cref{sec:boltzmann-sat-intro}, we designed a family of sequential Boltzmann distributions with the goal of investigating whether lookahead is beneficial for autoregressive models. We give the specifics of our setting below.

\paragraph{Dataset} In our experiments, we set the number of variables to be $n=15$. 
To keep the clause density close to the threshold, we set the number of clauses to be $m=64$ so that $\frac{m}{n}=4.267\approx \widehat{\alpha}_3\approx 4.269$.
We first sample 3-\sat formulas from the random 3-\sat distribution as in \cref{sec:boltzmann-sat-intro}. Each sampled formula yields a completely different instance of the task.  For each sampled formula, we derive a probability distribution, sample training and test data, fit a model of the distribution on training data, and evaluate the model on test data.  We compare the performance of different model architectures across all of the sampled formulas.

Concretely, for each formula, we define a Boltzmann distribution at temperature $T=2/3$, as described in \cref{sec:boltzmann-sat-intro}. Because $n$ is relatively small, we can explicitly compute the probabilities of all $2^{15}=32768$ assignments, and then explicitly compute all of the autoregressive conditional probabilities $p(x_t \mid \vx_{<t})$. Given any assignment $\vx$ of the variables to this formula, e.g., $\vx=\texttt{1 0 0 1 1 0 0 1 0 1 1 1 0 1 1}$, we task the model to learn the autoregressive conditional distributions.  More precisely, we take the first 5 bits $\vx_{1:5}$ as a prompt $\vec{x}^\text{source}$, and ask the model to predict $\vec{x}^\text{target} = \vx_{6:15}$.

To evaluate generalization, we split each dataset so that any prompt that appears in test data was never seen in training data.  In other words, we group the dataset of 32768 examples by the $2^5=32$ possible prefixes, so that each group contains $2^{10}=1024$ strings, and then randomly select 24 groups for training, 4 for validation, and 4 for test.  Then, for each dataset, the training, validation and test partitions consist of 24.6K, 4.1K, and 4.1K examples respectively.

\paragraph{Model and Training} Since we have a binary vocabulary in this setting, we use smaller embedding dimensions. We set $d_\text{model}=16$, $d_\text{FFN}=32$ and $n_\text{head}=2$ for all models. We train a Lookahead Transformer with 3 causal layers and 1 lookahead layer for each formula. For baseline comparison, we additionally train 3-layer, 4-layer and 5-layer Transformers.

We train our models with cross-entropy losses.
Let $p(x_{t+1}\mid \vx_{1:t})$ denote the exact conditional distribution of the Boltzmann distribution---which we compute by brute force as mentioned above---and let $\widehat{p}_\theta(x_{t+1}\mid \vx_{1:t})$ denote the estimated conditional distribution by our model.  
The loss over an assignment string $\vx$ is
\begin{align}
    \calL(\vx) = \sum_{t\geq5, i\in\{\texttt{0},\texttt{1}\}}
       & -p(x_{t+1}=i\mid \vx_{1:t}=\va_{1:t})  \nonumber \\
    & \mbox{} \cdot \log \widehat{p}_\theta(x_{t+1}=i\mid \vx_{1:t}=\va_{1:t})
\end{align}

We train the baseline Transformer models for 100 epochs and the Lookahead Transformers for 20 epochs. We use a learning rate of 0.02 for all models.

\paragraph{Results} We sampled a total of 50 random formulas (and hence trained 50 models for each setting, one for each formula). In \cref{tab:sat-results}, we report the loss evaluated on the test subset averaged over all 50 random formulas. We observe that, under our experimental setting, an additional lookahead layer has same amount of performance improvement as adding 2 additional causal layers.

\subsection{Multiple Letter Infilling} \label{sec:letter-infill-results}

\paragraph{Data Collection}
To facilitate the experiment as described in \cref{sec:letter-infill-intro}, we generate the dataset using the Unix standard \verb|words| file (located in \verb|/usr/share/dict/words|) and
independently choose whether to mask each character ($p=0.4$).
To ensure the regularity of the data source, we selected only alphabetic words (excluding those with digits) with a length ranging between 5 and 15 characters. The filtered subset constitutes 94.2\% of the \verb|words| file. We randomly partitioned our collected dataset into training, validation, and tests sets of of 201K, 10K, and 10K  examples respectively. An example sequence from this dataset is $\vec{x}^\text{source}=\verb|d - s - - e - a - c -|$ and $\vec{x}^\text{target}=\verb|d i s c r e p a n c y|$.

\paragraph{Model and Training} This dataset has a vocabulary size of $29 = 26 + |\{ \bos, \texttt{-}, \eos \}|$, where \bos and \eos mark the beginning and end of the target sequence.  Accordingly,  we set $d_\text{model}=24,d_\text{ffn}=96,n_\text{head}=4$ across all models for this dataset.
Given a number of base layers $L$ where $L=6$ or $10$, we train Lookahead Transformers with 1 and 2 additional lookahead layers and compare them against ordinary baseline Transformers of $L$, $L+1$ and $L+2$ layers. All baseline models are trained for 200 epochs whereas the lookahead models are trained for 40 epochs. 
For the 6, 7 and 8 layer baseline Transformer as well as the 6+1 and 6+2 Lookahead Transformers, we use a learning rate of 5e-3. For all the other Transformers with $\geq$10 layers, we use a learning rate of 2.5e-3.

\paragraph{Results} 
We report both test set loss and accuracy for all our models in \cref{fig:synth-test-results}. 
Due to space limits, the validation set statistics can be found in \cref{fig:synth-valid-results} in \cref{sec:valid-results}. 
We again observe that, in this task, adding a single lookahead layer has similar or more performance improvement as adding two additional standard causal layers. For example, a 10+1 Lookahead Transformer has lower loss and higher accuracy than a 12-layer baseline Transformer, even though the latter has more parameters.
In this particular task, in the regime where the baseline model is almost plateaued in terms of improvement given an additional layer (i.e., when it has 10, 11 and 12 layers), adding a lookahead layer is particularly helpful.

The learning curves are shown in \cref{sec:learning-curves}. We note from \cref{fig:learning-curves} that, almost immediately after the training of the lookahead models begins, they are able to improve upon the baseline.

\subsection{Morphological Inflection} \label{sec:morph-results}

\paragraph{Data Processing} We use the medium data from task 1 of the CoNLL-SIGMORPHON 2017 shared task \citep{cotterell-etal-2017-conll}, which contains 1000 examples per language over 52 languages. We adopt the multilingual training paradigm commonly used in morphological inflection — instead of training a separate model for each language, we train a single joint model of all the languages at the same time, allowing parameters to be shared between languages \citep{bergmanis-etal-2017-training}. As such, each sample contains a language identification token, the morphological tags, the lemma and the inflected form. An example would be $\vec{x}^\text{source}=\verb|english s p a r k verb participle past|$, $\vec{x}^\text{target}=\texttt{s p a r k i n g}$.

\paragraph{Model and Training} Despite the much larger vocabulary size (759) of all characters of 52 languages plus special tokens, we found that using the same dimensions as in our letter infilling experiment (\cref{sec:letter-infill-results}) works very well in practice. 
Therefore, following the previous experiment, we again set $d_\text{model}=24,d_\text{ffn}=96,n_\text{head}=4$ across all models for this dataset. 
We also keep the experimental setting where we train the Lookahead Transformer with 6 or 10 base layers and 1 or 2 lookahead layers and compare against baseline Transformers with matching number of parameters. Similar to \cref{sec:letter-infill-results}, we train the baseline models for 200 epochs and the lookahead models for 40 epochs. We also use the same setting of learning rates as in \cref{sec:letter-infill-results}.

\paragraph{Results} 
We report both loss and accuracy for all models in \cref{fig:morph-test-results} and \cref{fig:morph-valid-results}. We first observe the same pattern where the improvement of one additional lookahead layer is comparable to two causal layers. We also notice that, in this dataset, there is much less additional improvement if we add one more lookahead layer. We observe similar effect in \cref{sec:letter-infill-results}. These indicate that the biggest improvement comes from performing rollouts, whereas having more layers to analyze the lookahead strings have diminishing returns.

\begin{figure*}[t]
    \centering
    \includegraphics[width=\textwidth]{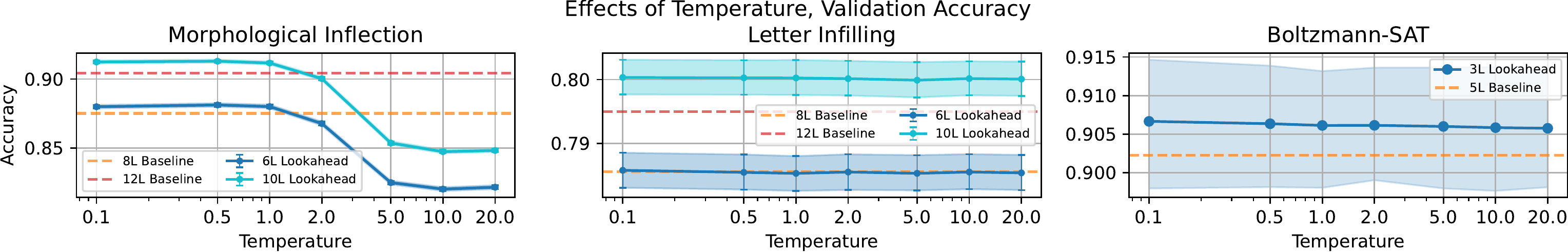}
    \caption{Validation accuracy of 6 and 10 layer Transformer with 1 layer of lookahead under proposals adjusted with varying temperature ($95\%$ error bars computed using bootstrap resampling).
    The Lookahead Transformers are all trained with the original proposal distribution. 
    Figure shows results with the proposal adjusted during inference time.}
    \label{fig:ablation-temperature-accuracy-results}
\end{figure*}

\begin{figure*}[t]
    \centering
    \includegraphics[width=\textwidth]{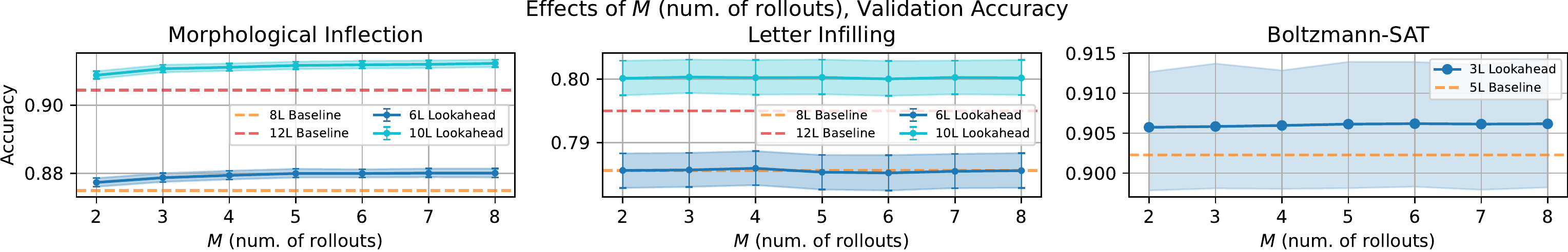}
    \includegraphics[width=\textwidth]{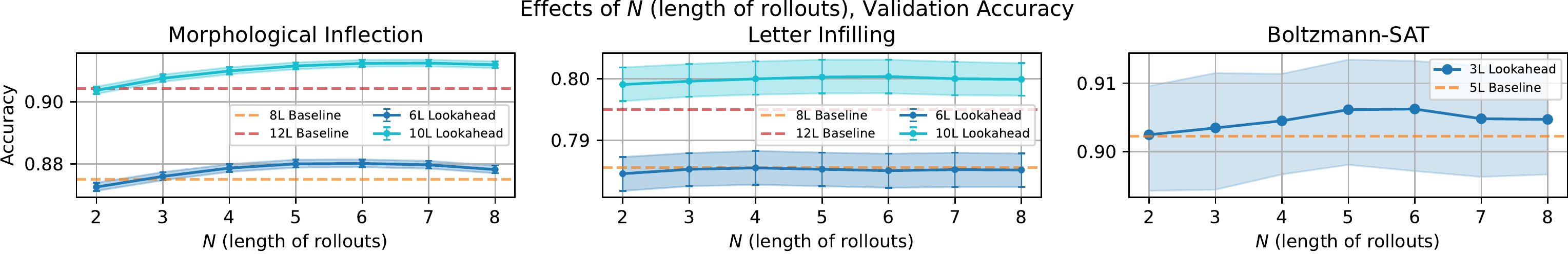}
    \caption{Validation accuracy of 6 and 10 layer Transformer with 1 layer of lookahead and varying numbers ($M$) and lengths ($N$) of rollouts ($95\%$ error bars computed using bootstrap resampling).
    The Lookahead Transformers are all trained with $N=M=5$.
    Figure shows results with varying $N$ and $M$ during inference time.
    }
    \label{fig:ablation-n-futures-accuracy-results}
\end{figure*}

\section{Ablation Studies}
\label{sec:ablation}

In this section, we study the effects of several parameters in the lookahead model.

\subsection{Effects of Number and Length of Rollouts}\label{sec:rollouts-ablation}
First, we aim to study the effects of varying the number ($M$) or length ($N$) of the rollouts. 
To this end, we evaluate the lookahead models with 1 lookahead layer under different $M$ and $N$ parameters.\footnote{Note that these models are still trained with $M=5, N=5$.} The accuracy statistics are shown in \cref{fig:ablation-n-futures-accuracy-results} and the loss statistics are shown in \cref{fig:ablation-n-futures-loss-results} in the appendix. 

In general, having fewer or shorter rollouts does make the lookahead models slightly worse, but they still outperform or are similar to baseline models with more parameters. This suggests that we may salvage most of the information contained in the rollouts by looking ahead a few steps.

\subsection{Effects of the Quality of Proposal Distribution}\label{sec:temperature}

Next, we are interested in the question of how much the lookahead model depends on the proposal distribution $q(\cdot)$. We answer this question by evaluating the lookahead models with proposal distributions modified with a temperature parameter $\tau$:
\begin{equation}
    q_\tau(x_{t+i,m}=v\mid\cdot)\propto q(x_{t+i,m}=v\mid\cdot)^{\frac{1}{\tau}}
\end{equation}
as an attempt to control the quality of the proposal distribution.  Intuitively, a high temperature ($\tau\gg 1$) means that the proposal is reduced to uniformly random rollouts, carrying no educated guess at all to the lookahead model, whereas a low temperature ($0<\tau\ll 1$) corresponds to a very sharp proposal distribution in which all of the rollouts tend to be the same string (the most probable string under $p_0$).
Again, we only test lookahead models that have 1 lookahead layer.

The accuracy statistics are shown in \cref{fig:ablation-temperature-accuracy-results} and the loss statistics are shown in \cref{fig:ablation-temperature-loss-results} in the appendix.  

No model is hurt by lowering the temperature.  Perhaps this is not surprising, since lowering the temperature has the effect of reducing the effective sample size---which did not hurt in \cref{sec:rollouts-ablation}---while also making the method closer to deterministic and thus perhaps easier to train.  

What is more surprising is that raising the temperature---even to a very high value so that the lookahead strings are essentially random---only slightly hurts the letter infilling and Boltzmann-\textsc{sat} tasks.  This suggests that in these tasks, the trained lookahead models are not carefully examining the lookahead strings.  Perhaps they achieve their gains (over baseline models with more parameters) by learning how to use the extra embedding vectors for additional computation.  By contrast, in the morphological inflection task, the trained lookahead models do degrade rapidly as temperature increases, suggesting that in this case, the lookahead strings do matter.  We plan to study the lookahead strings to better understand the difference among tasks.

\section{Limitations and Future Work}
\label{sec:limitation-future-work}

\subsection{Computational Cost}

Our implementation of the method presented here results in a 60x slowdown, so the most important aspect of future work is to reduce computational cost.
Fortunately, there are many ways to speed up the method, some of which are supported by the results in the ablation studies.
\begin{itemize}
    \item{\textbf{Reducing length and/or number of rollouts.}}
    As indicated by the experimental results in \cref{sec:ablation}, reducing the length and/or number of rollouts could be a simple and promising solution.

    \item{\textbf{Reusing rollouts.}} 
    We can reuse the same rollouts over several steps.  That is, rollouts given $\sentence_{\leq t}$ could be used to predict not only $\token_{t+1}$ but also $\token_{t+2}$ and $\token_{t+3}$, say.  This is not unreasonable since all three would already have been predicted (jointly) from these rollouts given $\sentence_{\leq t}$ if the tokenization scheme had treated $\sentence_{t+1:t+3}$ as a single token. 

    \item{\textbf{Adaptive rollouts.}} 
    Our current method rolls out and embeds $M$ new futures at \emph{each} step $t$, but this may be overkill.
    We could instead consult a policy to choose $M$.  At ``easy'' time steps $t$ where the proposal distribution $q(\token_{t+1} \mid \sentence_{1:t})$ is accurate, the policy should tend to choose $M=0$, and we can then simply predict $q(\token_{t+1} \mid \sentence_{1:t})$.  Conversely, the policy should choose large $M$ at the ``difficult'' steps where lookahead will actually help.  Similarly, we can choose the rollout length $N$ from a policy.

    \item{\textbf{Distillation.}} 
    Lastly, we can distill the slow model (lookahead model) into a faster architecture (generic Transformer). This is an example of ``structure compilation'' \cite{daume08flat}.  The lookahead architecture should generalize well to situations outside the training dataset because it analyzes them in detail at test time; yet a sufficiently large generic Transformer should be able to capture the same patterns.  This Transformer may have too many parameters to learn directly from the original training dataset, but can be trained on a larger synthetic dataset generated from the lookahead model.

\end{itemize}

\subsection{Alternative Architectures and Approaches}

The bidirectional lookahead attention as described by \cref{eq:bidirect-attn} does not have the ability to directly distinguish between the lookahead strings and the prefix.
This design may partially account for the observation that a Lookahead Transformer often doesn't benefit from the content of the lookahead strings.
In future work, as a remedy, we can explicitly mark 
the lookahead tokens by adding an extra vector to them along with their positional embedding vectors. 
Alternatively, we can insert a special token to indicate the start of the lookahead string.

One could also consider quite different architectures for using lookahead.  Following the ``LM recursion'' ideas of \citet{levine-etal-2022-standing}, one could just autoregressively condition $\sentence_{t+1}$ on a prompt that is constructed from the prefix $\sentence_{\leq t}$ and all of the lookahead strings $\sentence_{>t}$.  The model used to conditionally predict $\sentence_{t+1}$ might simply be the base model $q$, or might be derived from $q$ using parameter-efficient fine-tuning.

Finally, in the broader sense, our Monte Carlo lookahead method is comparable to off-policy exploration in reinforcement learning.  In particular, Monte Carlo tree search \cite{browne2012mcts-survey} performs many rollouts from a state in order to assess the values of different next actions.  However, in our method, the information gathered from exploration is incorporated into the next-action policy in an unrestricted learned way.  In future work, it would be worth experimenting with more structured ways to use the information.  For example, inspired by the motivation in the introduction, as well as by MCTS, we might attempt to learn a reward function that explicitly scores the lookahead strings.  \Citet{chaffin-etal-2022-ppl} have considered this framework when the reward function is already known.

\section{Conclusions}

We have presented a type of autoregressive generative sequence model that has the ability to perform lookahead when predicting the next symbol.  Although the set of \emph{all possible} futures has size that is exponential in the time horizon, we incur only a constant-time slowdown by generating and attending to a \emph{constant number} of random rollouts, which are meant to be representative of the full set of possible futures.

We found experimentally that an $L$-layer autoregressive model, when augmented with 1 lookahead layer that attends to $M=5$ random futures, can match and often beat the loss or accuracy of an ordinary $(L+2)$-layer autoregressive model---even though the latter has $\frac{L+2}{L+1}$ times as many parameters.

This pattern held across all three of our diverse tasks.  Thus, our exploratory study demonstrates that there can be some predictive value in shallow analysis of random futures predicted by the model given the observed past.  On the other hand, although all three tasks do benefit from the additional test-time computation (green and orange blocks in \cref{fig:model}), our ablation study in \cref{sec:temperature} found that only in one of the three tasks does this computation actually depend on the content of the random futures.  Thus, as discussed in \cref{sec:limitation-future-work}, we plan to investigate alternative architectures and training methods so that the model will benefit from lookahead strings more.

\bibliography{lookahead_icml}

\clearpage
\appendix
\appendixpage

\section{Additional Experimental Results}

\subsection{Validation Set Results} \label{sec:valid-results}

\begin{figure}[ht]
    \centering
    \includegraphics[width=0.48\textwidth]{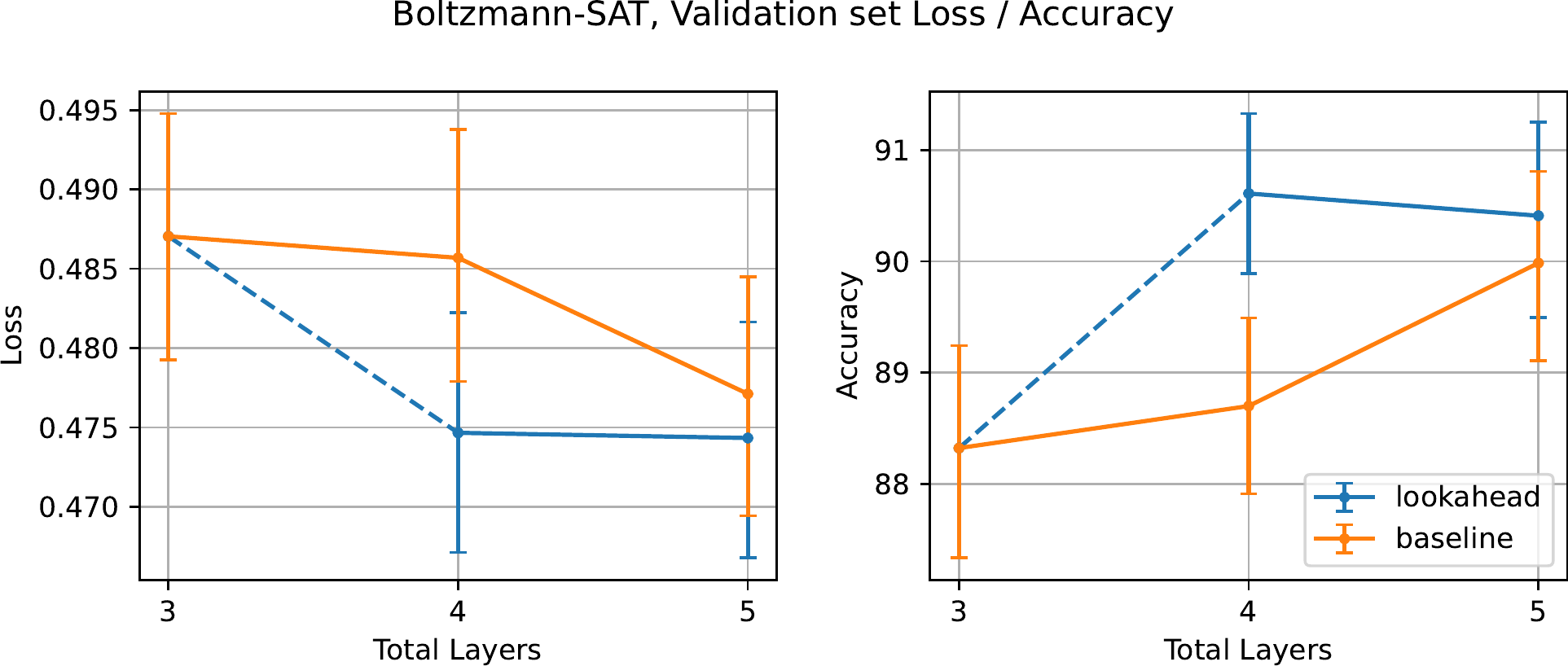}
    \caption{Mean loss over 50 formulas on validation set with error bars showing a 95\% bootstrap confidence interval.}
    \label{fig:sat-valid-results}
\end{figure}

\begin{figure*}[ht]
    \centering
    \includegraphics[width=0.49\textwidth]{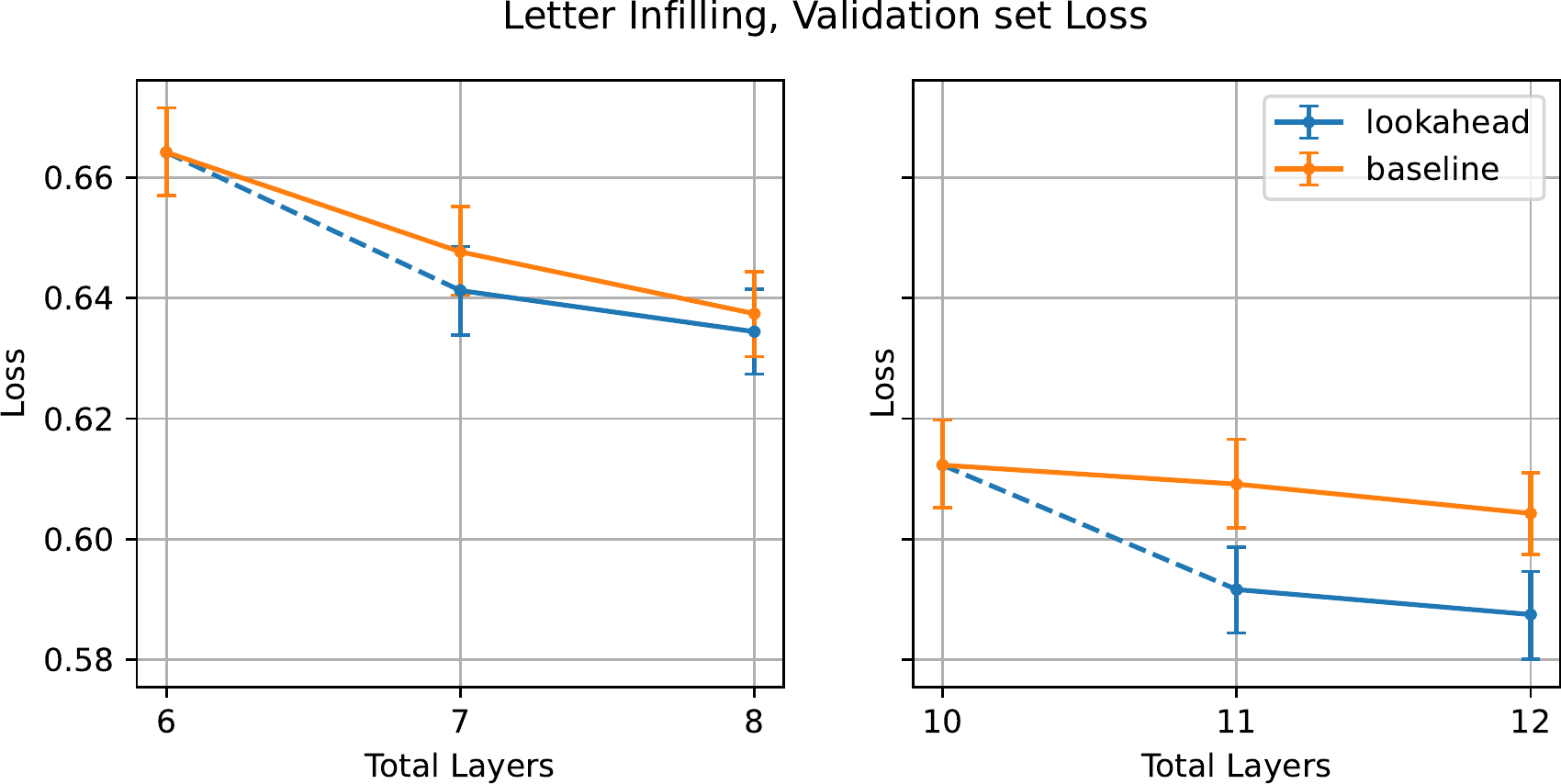}
    \includegraphics[width=0.49\textwidth]{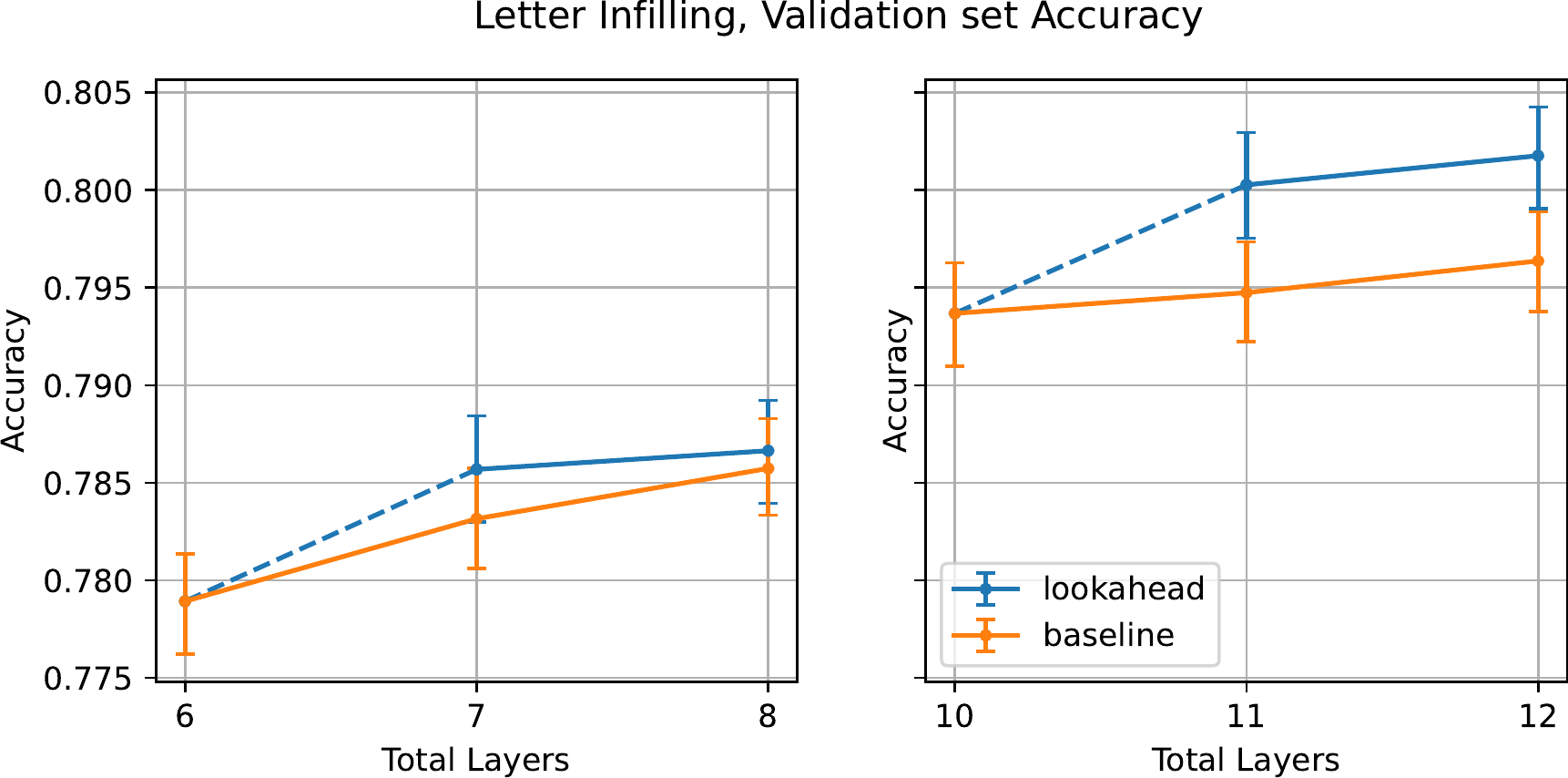}
    \caption{Validation set statistics of Lookahead Transformers and baseline Transformers with different number of layers for the letter infilling task (with $95\%$ error bars computed using bootstrap resampling). In each plot, the blue line shows the result of adding 0, 1, or 2 bidirectional lookahead attention layers to the 6 or 10 base causal layers, whereas the orange baseline shows the result of adding 0, 1, or 2 ordinary causal layers.  All points are trained for the same total number of epochs.
    }
    \label{fig:synth-valid-results}
\end{figure*}

\begin{figure*}[ht]
    \centering
    \includegraphics[width=0.49\textwidth]{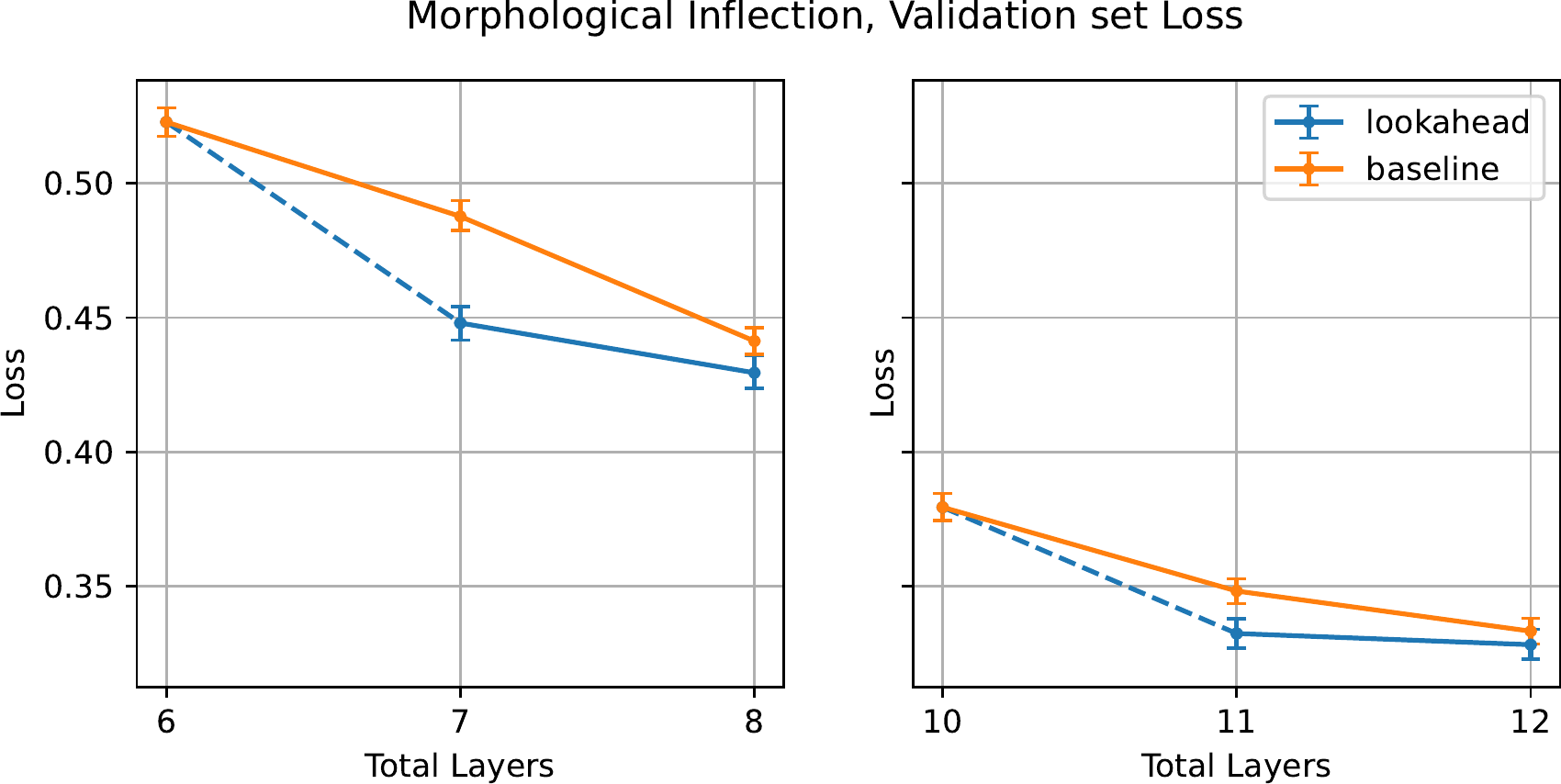}
    \includegraphics[width=0.49\textwidth]{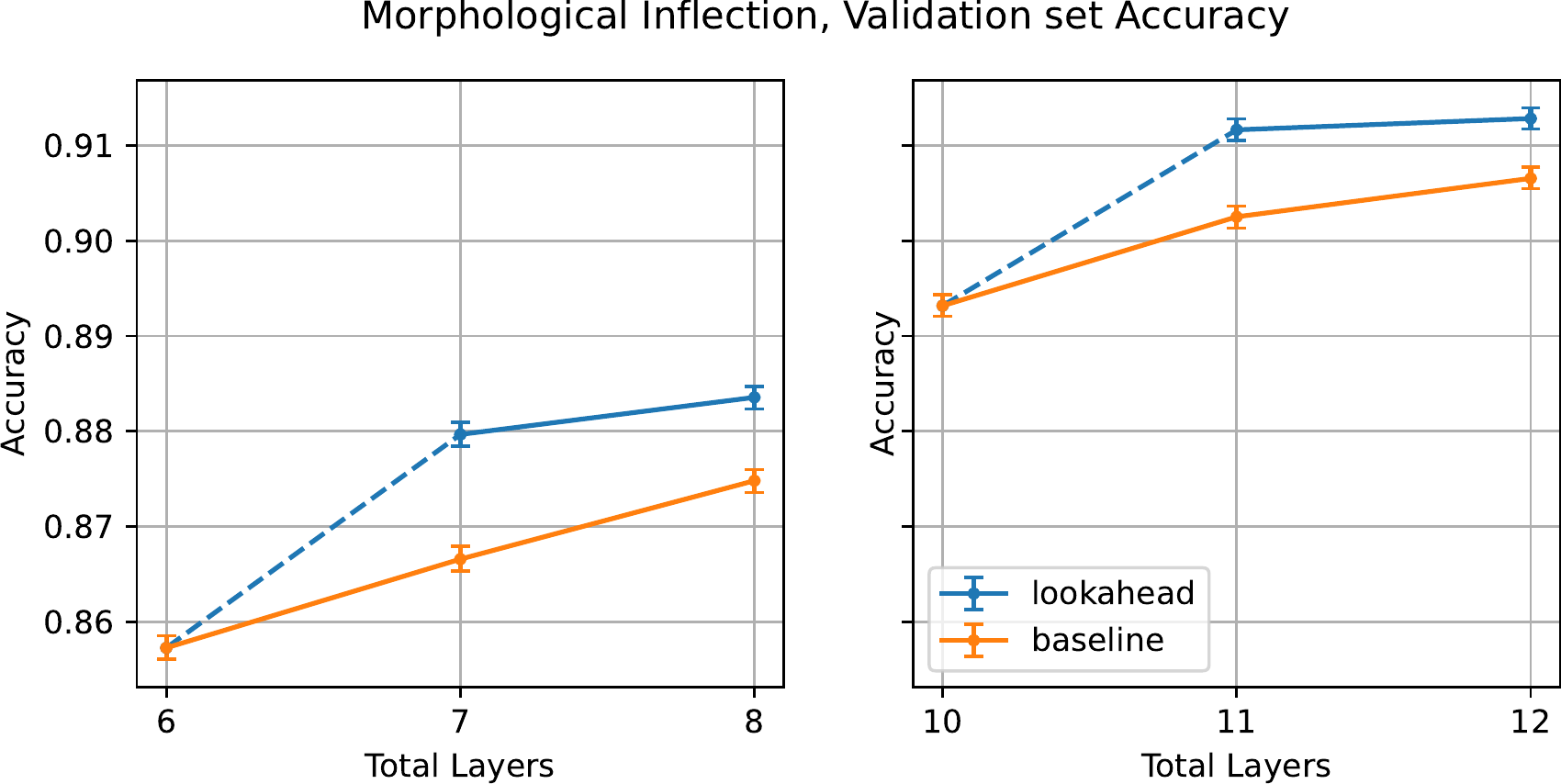}
    \caption{Validation set statistics of Lookahead Transformers and baseline Transformers with different number of layers for morphological inflection (with $95\%$ error bars computed using bootstrapping). In each plot, the Lookahead Transformer has either 6 or 10 base causal layers, and the other one or two layers are the bidirectional lookahead attention layers.}
    \label{fig:morph-valid-results}
\end{figure*}

We have presented the test set results for learning the Boltzmann-\textsc{sat} distribution in \cref{sec:sat-results}, letter infilling in \cref{sec:letter-infill-results} and morphological inflection in \cref{sec:morph-results}.
Here, we present the results on validation sets for these tasks in \cref{fig:sat-valid-results}, \cref{tab:sat-valid-results}, \cref{fig:synth-valid-results} and \cref{fig:morph-valid-results}.

\begin{table}[ht]
    \centering
    \begin{tabular}{lcc}
    \toprule
         & Avg. Loss ($\downarrow$) & Avg. Acc ($\uparrow$) \\
    \midrule
       3-Layer Baseline  & 0.487 & 88.3 \\
       4-Layer Baseline  & 0.486 & 88.7 \\
       5-Layer Baseline  & {\bf 0.477} & {\bf 90.0} \\
    \midrule
       (3+1)-Layer Lookahead & {\bf 0.475} & {\bf 90.6} \\
       (3+2)-layer Lookahead & {\bf 0.475} & {\bf 90.4} \\
    \bottomrule
    \end{tabular}
    \caption{Validation loss (log-loss per token) and argmax-prediction accuracy averaged over 50 randomly sampled 3-\sat formulas for Lookahead Transformer and baseline Transformers.  We boldface the best (lowest) result and all others that are not significantly worse (paired permutation test by formula, $p<0.05$).}
    \label{tab:sat-valid-results}
\end{table}

\begin{figure*}[ht]
    \centering
    \includegraphics[width=0.49\textwidth]{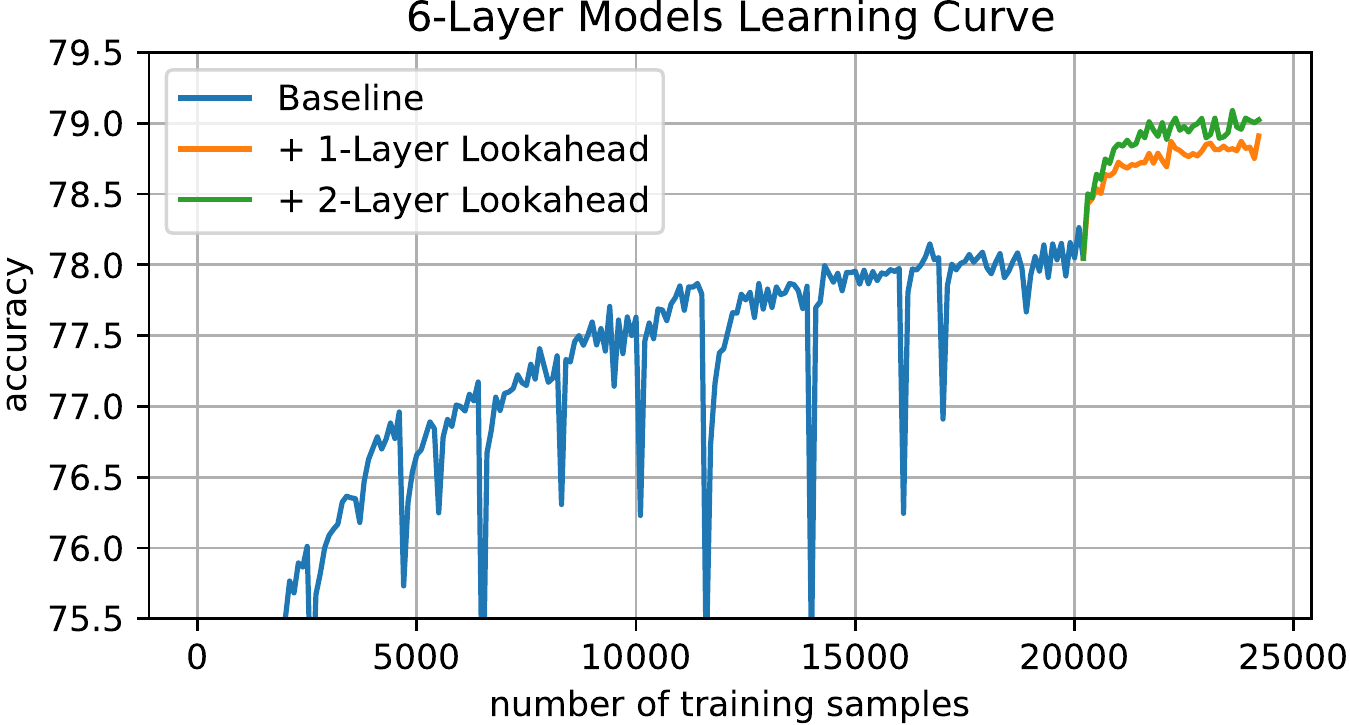}
    \includegraphics[width=0.49\textwidth]{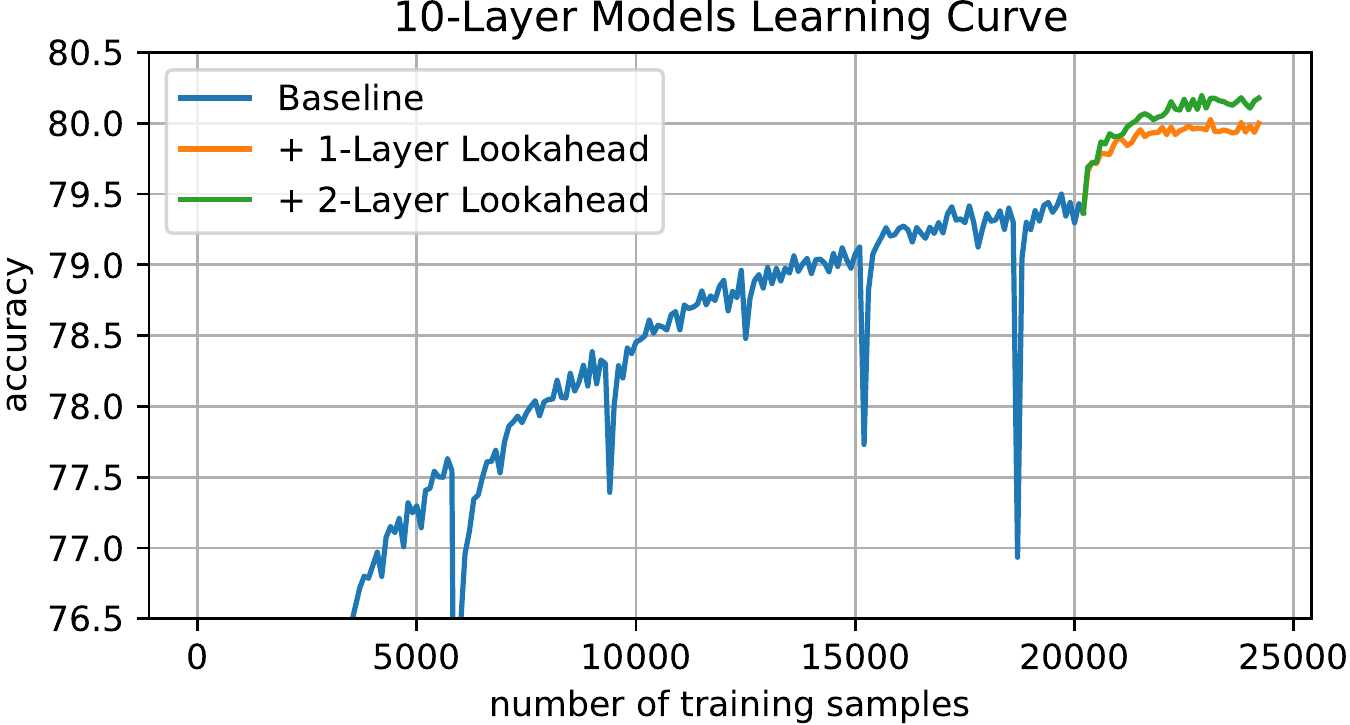}
    \includegraphics[width=0.49\textwidth]{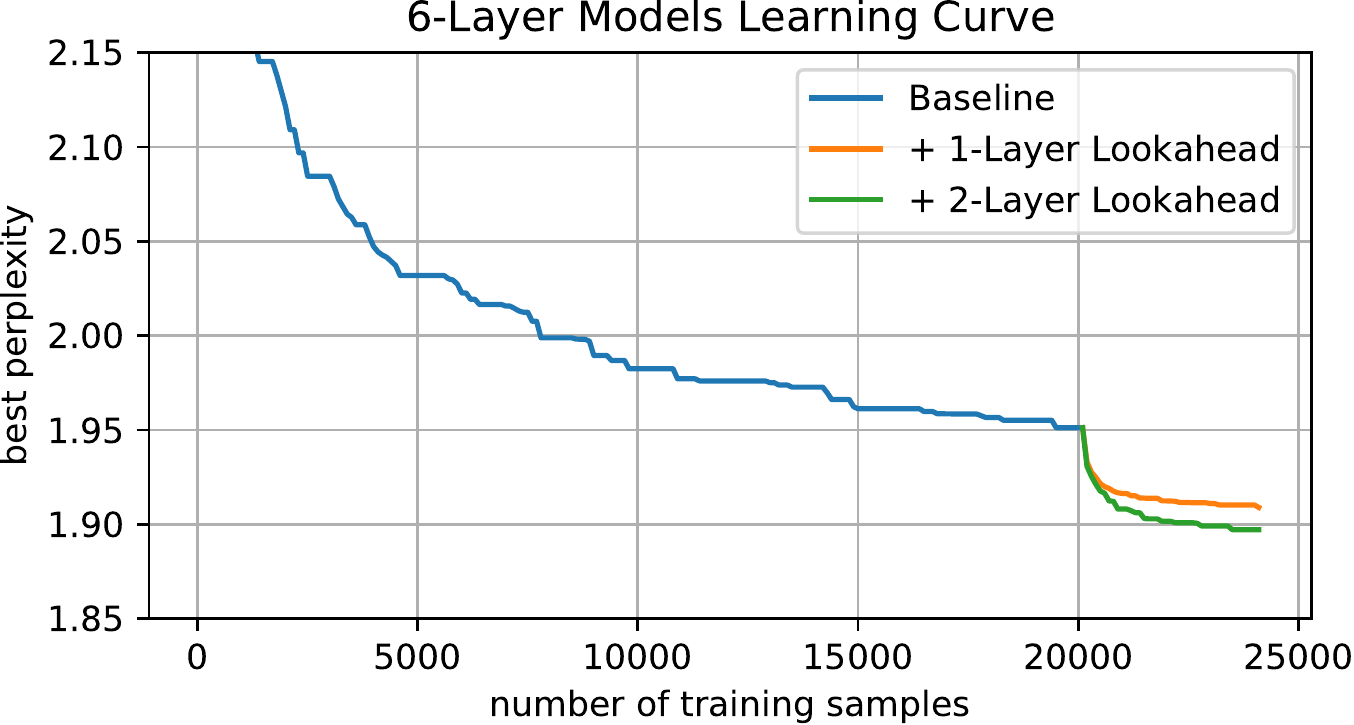}
    \includegraphics[width=0.49\textwidth]{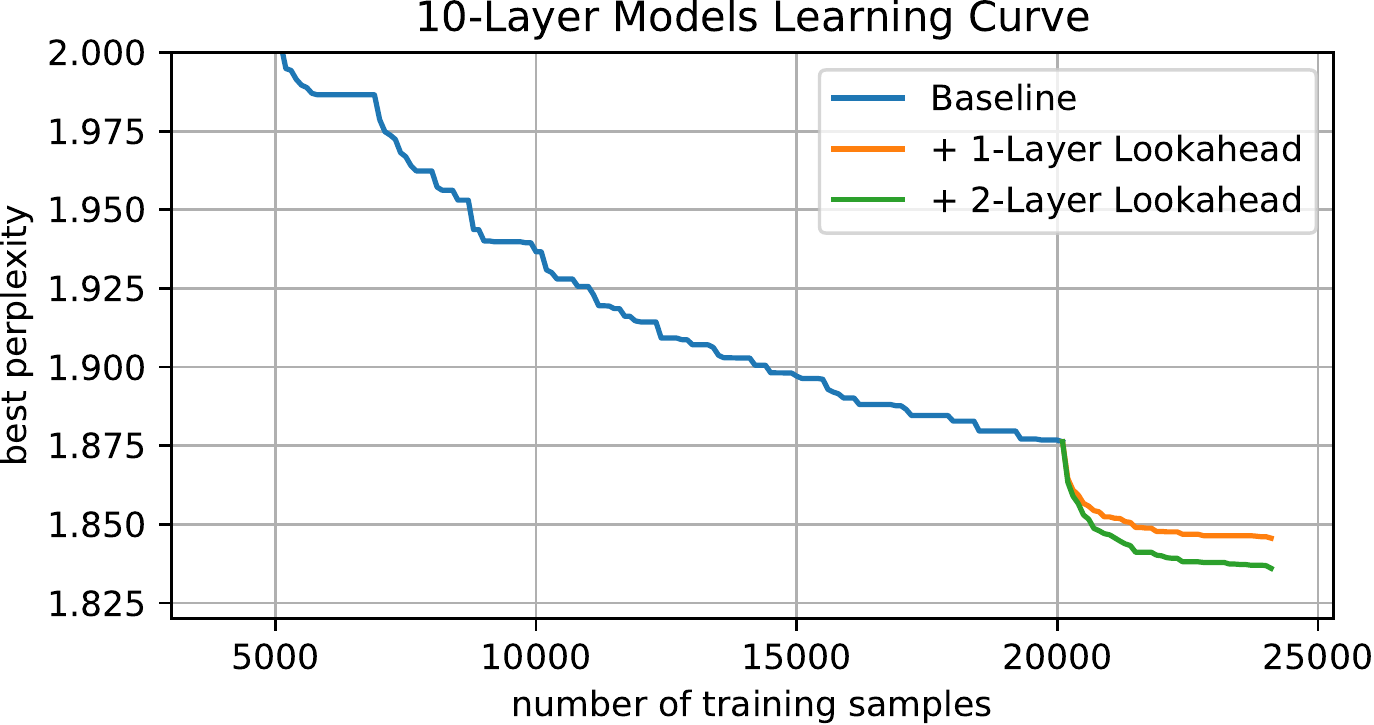}
    \caption{Learning curves on the validation set for the letter infilling task.}
    \label{fig:learning-curves}
\end{figure*}

\subsection{Learning Curves} \label{sec:learning-curves}
\cref{fig:learning-curves} shows the learning curves of two different metrics on the validation set for the letter infilling task. We observe that additional lookahead layers improves upon the baseline soon after the training begins. We also note we observed this pattern in early experiments and it holds across all our experiments.

\section{Ablation Study Details}

\begin{figure*}[t]
    \centering
    \includegraphics[width=\textwidth]{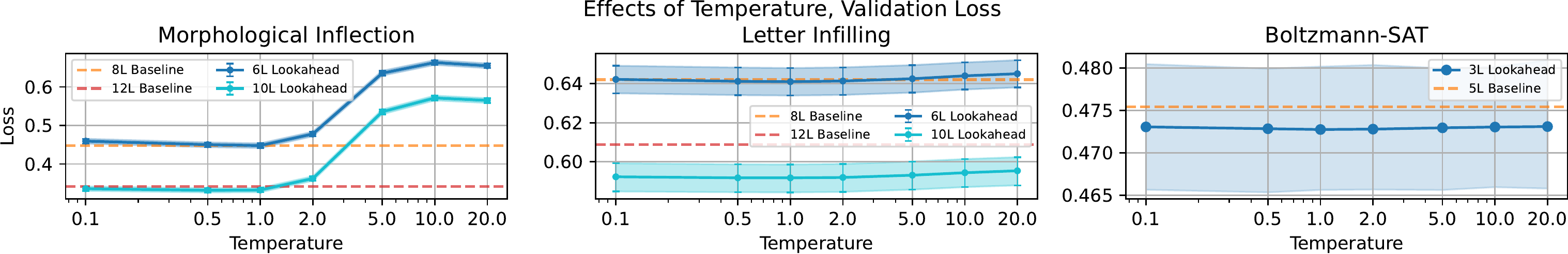}
    \caption{Validation loss of 6 and 10 layer Transformer with 1 layer of lookahead under proposals adjusted with varying temperature ($95\%$ error bars computed using bootstrap resampling).
    The Lookahead Transformers are all trained with the original proposal distribution. 
    Figure shows results with the proposal adjusted during inference time.}
    \label{fig:ablation-temperature-loss-results}
\end{figure*}

\begin{figure*}[t]
    \centering
    \includegraphics[width=\textwidth]{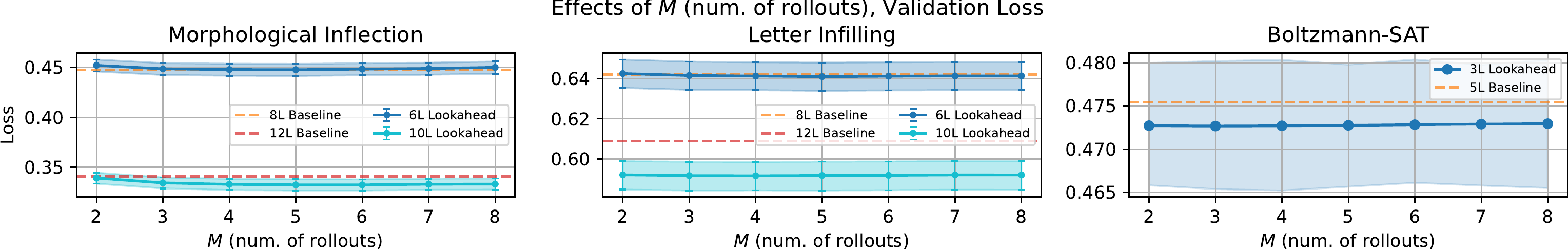}
    \includegraphics[width=\textwidth]{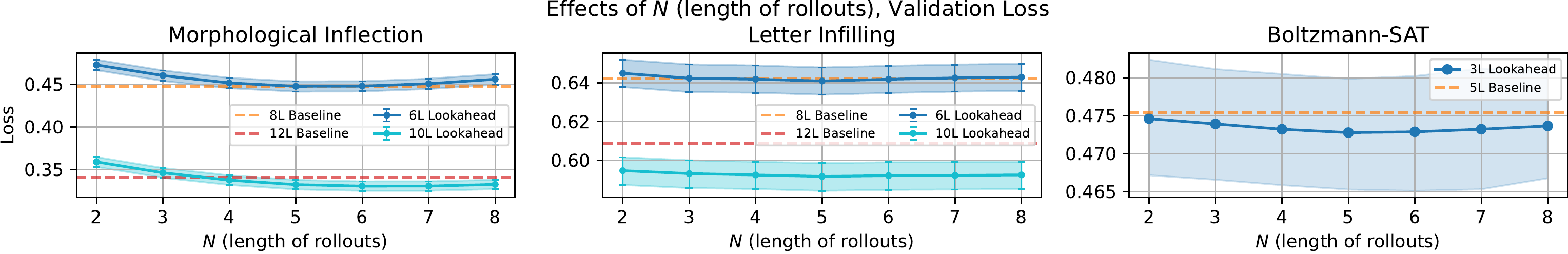}
    \caption{Validation loss of 6 and 10 layer Transformer with 1 layer of lookahead and varying numbers ($M$) and lengths ($N$) of rollouts ($95\%$ error bars computed using bootstrap resampling).
    The Lookahead Transformers are all trained with $N=M=5$.
    Figure shows results with varying $N$ and $M$ during inference time.}
    \label{fig:ablation-n-futures-loss-results}
\end{figure*}

We report the loss statistics on validation sets for ablation studies here. The accuracy statistics are shown in the main paper. The loss statistics are reported in \cref{fig:ablation-temperature-loss-results} and \cref{fig:ablation-n-futures-loss-results}.

\end{document}